\documentclass[10pt,journal,compsoc]{IEEEtran}
\usepackage{times}

\usepackage{soul}
\usepackage{url}
\usepackage[hidelinks]{hyperref}
\usepackage[utf8]{inputenc}
\usepackage[small]{caption}
\usepackage{graphicx}
\usepackage{amsmath}
\usepackage{booktabs}
\usepackage{multirow}
\usepackage{multicol}
\usepackage{amsmath}
\usepackage{amsthm}
\usepackage{amsfonts}
\usepackage{booktabs}
\usepackage{algorithm}
\usepackage{algpseudocode}

\newcommand{\nosection}[1]{\vspace{2pt}\noindent\textbf{#1.}}
\ifCLASSOPTIONcompsoc
  \usepackage[nocompress]{cite}
\else
  \usepackage{cite}
\fi
\ifCLASSINFOpdf
\else
\fi
\begin{document}

\title{A Survey of Adversarial Learning on Graphs}

\author{Liang Chen, Jintang Li, Jiaying Peng, Tao Xie, \\Zengxu Cao, Kun Xu, Xiangnan He, Zibin Zheng, Bingzhe Wu 
\IEEEcompsocitemizethanks{
\IEEEcompsocthanksitem Liang Chen, Jintang Li, Jiaying Peng, Tao Xie, Kun Xu, and Zibin Zheng are with Sun Yat-sen University. E-mail: \{chenliang6,zhzibin\} @mail.sysu.edu.cn, \{lijt55,pengjy36,xiet23\}@mail2.sysu.edu.cn
\IEEEcompsocthanksitem Zengxu Cao is with Hangzhou Dianzi University of China. E-mail: czx@hdu.edu.cn
\IEEEcompsocthanksitem Xiangnan He is with University of Science and Technology of China. E-mail: xiangnanhe@gmail.com
\IEEEcompsocthanksitem Bingze Wu is with Tencent AI Lab.  E-mail: wubingzhe94@gmail.com}
\thanks{Corresponding to Liang Chen and Zibin Zheng.}}


\markboth{Preprint: A Survey of Adversarial Learning on Graphs}%
{Shell \MakeLowercase{\textit{et al.}}: Bare Demo of IEEEtran.cls for Computer Society Journals}

\IEEEtitleabstractindextext{%
\begin{abstract}
	Deep learning models on graphs have achieved remarkable performance in various graph analysis tasks, e.g., node classification, link prediction, and graph clustering. However, they expose uncertainty and unreliability against the well-designed inputs, i.e., \emph{adversarial examples}.  Accordingly, a line of studies has emerged for both attack and defense addressed in different graph analysis tasks, leading to the arms race in graph adversarial learning. Despite the booming works, there still lacks a unified problem definition and a comprehensive review. To bridge this gap, we investigate and summarize the existing works on graph adversarial learning tasks systemically. Specifically, we survey and unify the existing works w.r.t. attack and defense in graph analysis tasks, and give appropriate definitions and taxonomies at the same time.  Besides, we emphasize the importance of related evaluation metrics, investigate and summarize them comprehensively. Hopefully, our works can provide a comprehensive overview and offer insights for the relevant researchers. Latest advances in graph adversarial learning are summarized in our GitHub repository \href{https://github.com/EdisonLeeeee/Graph-Adversarial-Learning}{https://github.com/EdisonLeeeee/Graph-Adversarial-Learning}.
\end{abstract}

\begin{IEEEkeywords}
Adversarial Learning, Graph Neural Networks, Adversarial Attack and Defense, Adversarial Example
\end{IEEEkeywords}}

\maketitle

\IEEEdisplaynontitleabstractindextext

%
\IEEEpeerreviewmaketitle

\IEEEraisesectionheading{\section{Introduction}}
\IEEEPARstart{O}{ver} the past decade, deep learning has enjoyed the status of crown jewels in artificial intelligence, which shows an impressive performance in various applications, including speech and language processing \cite{xiong2016achieving,collobert2011natural}, face recognition \cite{parkhi2015deep} and object detection \cite{krishnamurthy2018deep}. However, the frequently used deep learning models recently have been proved unstable and unreliable due to the vulnerability against perturbations. For example, slight changes on several pixels of a picture, which appears imperceptible for human eyes but strongly affect the outputs of deep learning models \cite{papernot2017practical}. As stated by Szegedy et al. \cite{szegedy2013intriguing}, deep learning models that are well-defined and learned by backpropagation have intrinsic blind spots and non-intuitive characteristics,  which should have been generalized to the data distribution in an obvious way.

\begin{figure}
	\centering
	\includegraphics[width=\linewidth]{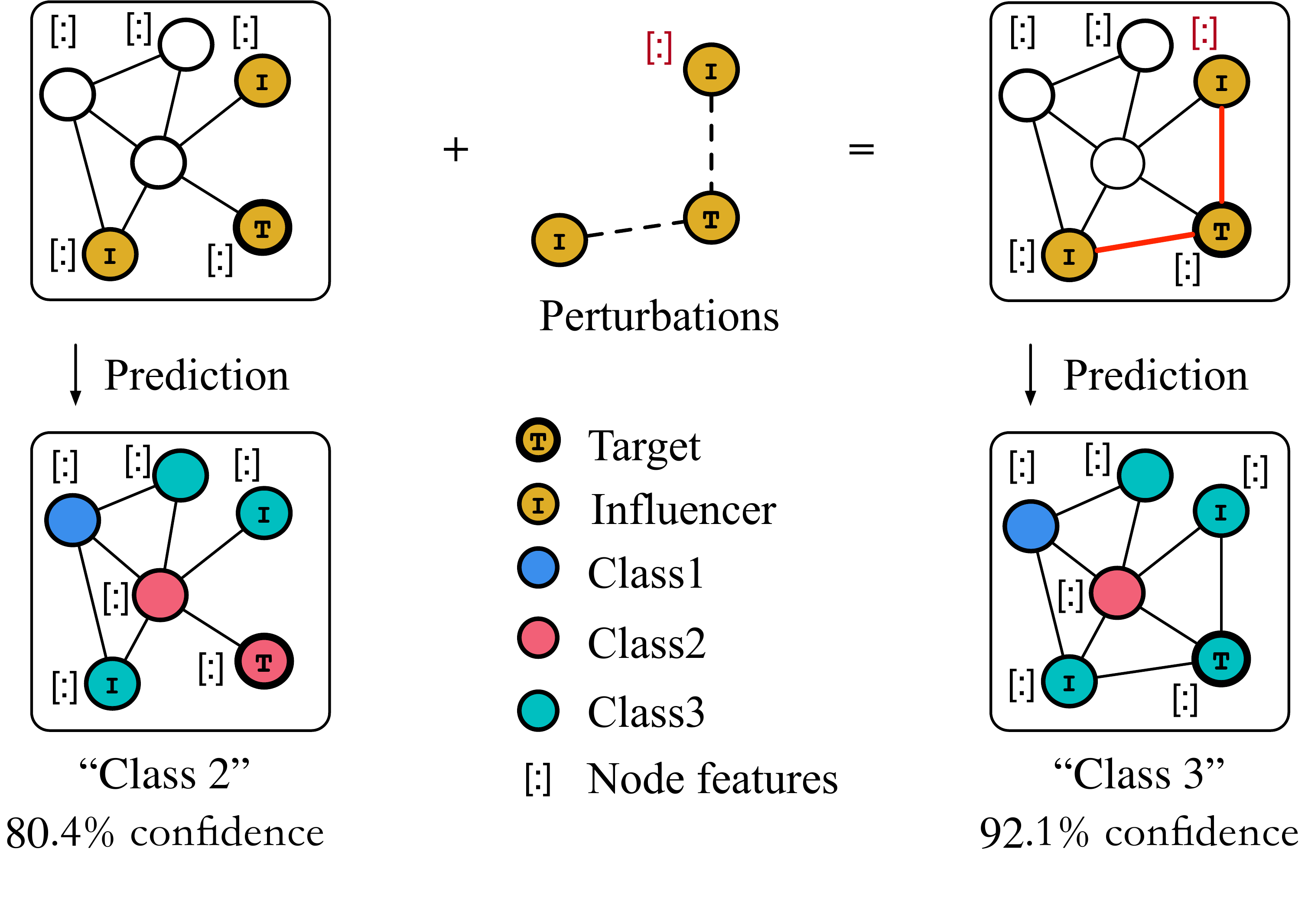}\\
	\caption{A misclassification of the target caused by a small perturbations of the graph structure and node features.}
	\label{Fig1}
\end{figure}

On the other hand, deep learning on graphs has received significant research interest recently. As a powerful representation, graph plays an important role and has been widely applied in real world \cite{goyal2018graph}. Naturally, deep learning's research on graph is also a hot topic and brings lots of refreshing implementations in different fields, such as social networks \cite{pei2015nonnegative}, e-commence networks \cite{wang2018billion} and recommendation systems \cite{chen2018heterogeneous,xie2018factorization}. Unfortunately, graph analysis domain, a crucial field of machine learning, has also exposed the vulnerability of deep learning models against well-designed attacks \cite{zugner2018adversarial,zugner2019certifiable}. For example, consider the task of node classification, attackers usually have control over several fake nodes, aim to fool the target classifiers, leading to misclassification by adding or removing edges between fake nodes and other benign ones. As shown in Figure \ref{Fig1}, performing small perturbations (two added links and several changed features of nodes) on a clean graph can lead to the misclassification of deep graph learning models.

With rising concerns being paid on the security of graph models, there exists a surge of researches on graph adversarial learning, i.e., a field on studying the security and vulnerability of graph models. On one hand, from the perspective of attacking a graph learning model, Z\"ugner et al. \cite{zugner2018adversarial} first study adversarial attacks on graph data, with little perturbations on node features and graph structure, the target classifiers are easily fooled and misclassify specified nodes. On the other hand, Wang et al. \cite{wang2019adversarial} propose a modified Graph Convolution Networks (GCNs) model with an adversarial defense framework to improve its robustness. Moreover, Sun et al. \cite{sun2018adversarial} study the existing works of adversarial attack and defense strategies on graph data and discuss their corresponding contributions and limitations. However, they mainly focus on the aspect of the adversarial attack, leaving works on defense unexplored.

\nosection{Challenges} Despite a surge of works on the graph adversarial learning, there still exists several problems to solve. i) \emph{Unified and specified formulation}. Current studies consider the problem definition and assumptions in graph adversarial learning with their own mathematical formulations and mostly lack of detailed explanations, which effectively hinders the progress of follow-up studies. ii) \emph{Related evaluation metrics}. While for various tasks, evaluation metrics on corresponding performance are rather different, and even have diverse standardization on it. Besides, special metrics for the graph adversarial learning scenario are necessary and timely to explore, e.g., evaluation on the attack impacts.

For the problem of inconsistent formulations and definitions, we survey the existing attack and defense works, give unified definitions and categorize them from diverse perspectives. Although there have been some efforts \cite{zugner2018adversarial,ma2019attacking,dai2018adversarial} to generalize the definitions, most formulations still make customization for their own models. So far, only one article \cite{sun2018adversarial} outlines these concepts from a review perspective, which is not sufficient to summarize the existing works comprehensively. Based on the previous literature, we summarize the different types of graphs and introduce the three main tasks according to the level, and subsequently give the unified formulations of attack and defense in Section \ref{attackDefinition} and \ref{defenseDefinition}, respectively.

Different models have various metrics due to different emphasis. To provide guidance for researchers and better evaluate their adversarial models, we have a more detailed summarize and discussion on metrics in Section \ref{metrics}. In particular, we first introduce some common metrics for both attack and defense, and then present some special metrics provided in their respective works from three categories: effectiveness, efficiency, and imperceptibility. For instance, the Attack Success Rate (ASR) \cite{chen2018link} and the Average Defense Rate (ADR) \cite{chen2019can} are proposed to measure the effectiveness of attack and defense, respectively.

In summary, our contributions can be listed as bellow:
\begin{itemize}
	\item We thoroughly investigate related works in this area, and subsequently give the unified problem formulations and the clear definitions for current inconsistent concepts of both attack and defense.
	\item We give a clear overview on existing works and classify them from different perspectives based on reasonable criteria systematically.
	\item We emphasize the importance of evaluation metrics and investigate them to make a comprehensively summary.
	\item For such an emerging research area, we point out the limitations of current researches and provide some open questions to solve.

\end{itemize}
The survey is organized as follow. In Section \ref{preliminary}, we will give some basic notations of typical graphs. In Section \ref{attack} and Section \ref{defense}, we will separately introduce the definitions, taxonomies of adversarial attack and defense on graph data, and further give a clear overview. We then summarize the related metrics in Section \ref{metrics} and try to discuss some open research questions in Section \ref{openProblems}. Finally, we draw our conclusion in Section \ref{conclusion}.

\begin{table*}[t]
	\centering
	\caption{Summary of notations.}
	\label{notation_table}
	\scalebox{0.8}{
		\begin{tabular}{c|c||c|c||c|c}
			\hline
			\hline
			Symbol                     &
			Description                &
			Symbol                     &
			Description                &
			Symbol                     &
			Description                  \\

			\hline
			\begin{tabular}[c]{@{}c@{}} $N$ \end{tabular}  &
			\begin{tabular}[c]{@{}c@{}} number of nodes \end{tabular}  &
			\begin{tabular}[c]{@{}c@{}} \\ $M$\\ \\ \end{tabular}  &
			\begin{tabular}[c]{@{}c@{}} number of edges \end{tabular}  &
			\begin{tabular}[c]{@{}c@{}} \\ $n$\\ \\ \end{tabular}  &
			\begin{tabular}[c]{@{}c@{}} number of graphs \end{tabular}
			\\

			\hline
			\begin{tabular}[c]{@{}c@{}} $F_{(\cdot)}$ \end{tabular} &
			\begin{tabular}[c]{@{}c@{}} number of features w.r.t. \\
				nodes $F_{(node)}$ or edges $F_{(edge)}$\end{tabular} &
			\begin{tabular}[c]{@{}c@{}} \\ $\tilde{G}$\\ \\ \end{tabular} &
			\begin{tabular}[c]{@{}c@{}} a graph instance, denoting \\
				a clean graph $G$ or a modified graph $\hat{G}$\end{tabular} &
			\begin{tabular}[c]{@{}c@{}} \\ $\mathcal{C}$\\ \\ \end{tabular} &
			\begin{tabular}[c]{@{}c@{}} set of pre-defined class labels \\ for nodes, edges or graphs \end{tabular}
			\\

			\hline
			\begin{tabular}[c]{@{}c@{}} \\ $\mathcal{V}$\\ \\ \end{tabular} &
			\begin{tabular}[c]{@{}c@{}} set of nodes $\mathcal{V}=\{v_1,v_2,\dots,v_N\}$ \end{tabular} &
			\begin{tabular}[c]{@{}c@{}} $\mathcal{E}$ \end{tabular} &
			\begin{tabular}[c]{@{}c@{}} set of edges $\mathcal{E}=\{e_1,e_2,\dots,e_M\}$ \end{tabular} &
			\begin{tabular}[c]{@{}c@{}} $\mathcal{G}$ \end{tabular} &
			\begin{tabular}[c]{@{}c@{}} set of graphs $\mathcal{G}=\{G_1,G_2,\dots,G_n\}$ \end{tabular}
			\\

			\hline
			\begin{tabular}[c]{@{}c@{}} $\mathcal{D}$ \end{tabular} &
			\begin{tabular}[c]{@{}c@{}} the whole dataset,\\ $\mathcal{D}=(\mathcal{G}, X, \mathcal{C})$ \end{tabular} &
			\begin{tabular}[c]{@{}c@{}} \\ $\mathcal{S}$\\ \\ \end{tabular} &
			\begin{tabular}[c]{@{}c@{}} set of instances, could be $\mathcal{G}$, $\mathcal{V}$, or $\mathcal{E}$ \end{tabular} &
			\begin{tabular}[c]{@{}c@{}} $\mathcal{S}_L$ \end{tabular} &
			\begin{tabular}[c]{@{}c@{}}  set of labeled instances, could be\\ $\mathcal{G}_L$, $\mathcal{V}_L$, or $\mathcal{E}_L$,  $S_L \subset S$ \end{tabular}
			\\

			\hline
			\begin{tabular}[c]{@{}c@{}} \\ $\mathcal{T}$\\ \\ \end{tabular} &
			\begin{tabular}[c]{@{}c@{}} set of unlabeled instances\\ where $\mathcal{T} \subset \mathcal{S}-\mathcal{S}_L$ or $\mathcal{T} = \mathcal{S}-\mathcal{S}_L$ \end{tabular} &
			\begin{tabular}[c]{@{}c@{}} \\ $\mathcal{K}$\\ \\ \end{tabular} &
			\begin{tabular}[c]{@{}c@{}}attackers' knowledge of the dataset \end{tabular} &
			\begin{tabular}[c]{@{}c@{}} \\ $\mathcal{O}$\\ \\ \end{tabular} &
			\begin{tabular}[c]{@{}c@{}}operation set\\ w.r.t. attackers' manipulation \end{tabular}
			\\

			\hline

			\begin{tabular}[c]{@{}c@{}}  $A$ \end{tabular} &
			\begin{tabular}[c]{@{}c@{}} adjacency matrix $A \in \mathbb{R}^{N\times N}$ \end{tabular} &
			\begin{tabular}[c]{@{}c@{}} \\ $A'$\\ \\ \end{tabular} &
			\begin{tabular}[c]{@{}c@{}} modified adjacency matrix \end{tabular} &
			\begin{tabular}[c]{@{}c@{}} $X$ \end{tabular} &
			\begin{tabular}[c]{@{}c@{}} feature matrix\\ $X \in \{0,1\}^{N \times F_{(\cdot)}}$ or $X \in \mathbb{R}^{N \times F_{(\cdot)}}$ \end{tabular}

			\\

			\hline

			\begin{tabular}[c]{@{}c@{}} \\ $X'$\\ \\ \end{tabular} &
			\begin{tabular}[c]{@{}c@{}} modified feature matrix \end{tabular} &
			\begin{tabular}[c]{@{}c@{}} $f$ \end{tabular} &
			\begin{tabular}[c]{@{}c@{}} deep learning model w.r.t. inductive \\ learning  $f^{(ind)}$ or transductive learning $f^{(tra)}$\end{tabular} &
			\begin{tabular}[c]{@{}c@{}} \\ $\hat{f}$\\ \\ \end{tabular} &
			\begin{tabular}[c]{@{}c@{}} surrogate model \end{tabular}

			\\

			\hline

			\begin{tabular}[c]{@{}c@{}} $\tilde{f}$ \end{tabular} &
			\begin{tabular}[c]{@{}c@{}} well-designed model for defense \end{tabular} &
			\begin{tabular}[c]{@{}c@{}} \\ $\theta$\\ \\ \end{tabular} &
			\begin{tabular}[c]{@{}c@{}} set of parameters\\ w.r.t. a specific model \end{tabular} &
			\begin{tabular}[c]{@{}c@{}} $Z$ \end{tabular} &
			\begin{tabular}[c]{@{}c@{}} output of the model \end{tabular}

			\\

			\hline
			\begin{tabular}[c]{@{}c@{}} \\ $t$\\ \\ \end{tabular} &
			\begin{tabular}[c]{@{}c@{}} target instance,\\ can be a node, an edge or a graph \end{tabular} &
			\begin{tabular}[c]{@{}c@{}} $\Delta$ \end{tabular} &
			\begin{tabular}[c]{@{}c@{}} attack budgets \end{tabular} &
			\begin{tabular}[c]{@{}c@{}} \\ $y$\\ \\ \end{tabular} &
			\begin{tabular}[c]{@{}c@{}} ground-truth label\\ w.r.t. an instance, $y \in \mathcal{C}$ \end{tabular}

			\\

			\hline
			\begin{tabular}[c]{@{}c@{}} \\ $\Psi$\\ \\ \end{tabular} &
			\begin{tabular}[c]{@{}c@{}} perturbation space \\of attacks \end{tabular} &
			\begin{tabular}[c]{@{}c@{}} \\ $\mathcal{L}$\\ \\ \end{tabular} &
			\begin{tabular}[c]{@{}c@{}} loss function of \\deep learning model \end{tabular} &
			\begin{tabular}[c]{@{}c@{}} \\ $\mathcal{Q}$\\ \\ \end{tabular} &
			\begin{tabular}[c]{@{}c@{}} similarity function \\of graphs\end{tabular}
			\\

			\hline
			\hline
		\end{tabular}
	}
\end{table*}

\section{Preliminary}
\label{preliminary}

Focusing on the graph structure data, we first give notations of typical graphs for simplicity and further introduce the mainstream tasks in graph analysis fields. The most frequently used symbols are summarized in Table \ref{notation_table}.
\subsection{Notations}
\label{notation}
Generally, a graph is represented as $\mathcal{G}=(\mathcal{V},\mathcal{E})$, where $\mathcal{V}=\{v_1, v_2, \dots, v_N \}$ denotes the set of $N$ nodes and $\mathcal{E}=\{e_1, e_2, \dots, e_M \}$ is the set of $M$ existing edges in the graph, and naturally $\mathcal{E} \subseteq \mathcal{V} \times \mathcal{V}$. The connections of the graph could be represented as an adjacency matrix $A \in \mathbb{R}^{N\times N}$, where $A_{i,j}\neq 0$ if there is an edge from node $v_i$ to node $v_j$ and $A_{i,j}=0$ otherwise.


\subsection{Taxonomies of Graphs}
Different scenarios correspond to various types of graphs, hence we will introduce them further in the following parts based on the basic graph definition in Section \ref{notation}.

\nosection{\textbf{Directed and Undirected Graph}} Directed graph, also called a digraph or a directed network, is a graph where all the edges are directed from one node to another  — but not backwards \cite{chartrand1977introductory}. On the contrary, a graph where the edges are bidirectional is called an undirected graph. For undirected graphs, the convention for denoting the adjacency matrix doesn't matter, as all edges are bidirectional. Generally, $A_{i,j}\neq A_{j,i}$ for directed graph and $A_{i,j}=A_{j,i}$ for undirected graph.

\nosection {\textbf{Weighted and Unweighted Graph}} Typically a weighted graph refers to an edge-weighted graph where each edge is associated with a real value \cite{chartrand1977introductory}. An unweighted graph may be used if a relationship in terms of magnitude doesn’t exist, i.e., the connections between edges are treated as the same.

\nosection {\textbf{Attributed Graph}}
An attributed graph refers to a graph where both node and edge are available to have its own attributes/features \cite{ding2019deep}. Specifically, the attributes of nodes and edges could be denoted as $X_{node} \in \mathbb{R}^{N\times F_{node}}$ and $X_{edge} \in \mathbb{R}^{M \times F_{edge}}$, respectively. In most cases a graph usually have attributes associated with nodes only, we use $X$ to denote the node attributes/features for brevity.

\nosection {\textbf{Homogeneous and Heterogeneous Information Graph}}
As well as attributes, the type of nodes or edges is another important property. A graph $\mathcal{G}$ is called heterogeneous information graph if there are two or more types of objects/nodes or relations/edges in it \cite{liu2018heterogeneous,hussein2018meta}; otherwise, it is called a homogeneous information graph.

\nosection {\textbf{Dynamic and Static Graph}}
Intuitively, nodes, edges and attributes are possibly changing over time in a dynamic graph \cite{manessi2020dynamic}, which could be represented at $\mathcal{G}^{(t)}$ at time $t$. For a static graph, which is simply defined as $\mathcal{G}$, in which nodes, edges and attributes remain the same as the time changes.

Each type of graph has different strengths and weaknesses. It's better to pick the appropriate kind of graph to model the problem. As existing works are mainly focus on a simple graph, i.e., undirected and unweighted. Besides, they assume that the graph is static and homogeneous for simplicity. By default, the mentioned ``graph'' refers to ``simple graph'' in this paper.

\subsection{Graph Analysis Tasks}
\label{task}
In this section, we will introduce the major tasks of graph analysis, in which deep learning models are commonly applied to. From the perspective of nodes, edges and graphs, we divide these tasks into three categories: node-, link- and graph-level task.

\nosection{\textbf{Node-level Task}}
Node classification is one of the most common node-level tasks, for instance, identifying a person in a social network. Given a graph $\mathcal{G}$, with partial labeled nodes $\mathcal{V}_L \subseteq \mathcal{V}$ and other unlabeled ones, the goal of classifiers is to learn a mapping function $\phi: \mathcal{V} \rightarrow \mathcal{C}$, where $\mathcal{C}$ is a set of pre-defined class labels. The learned mapping function is applied to effectively identify the class labels for the unlabeled nodes \cite{kipf2017semi}. To this end, inductive learning and transductive learning settings are specified based on the characteristic of training and testing procedures.
\begin{itemize}
	\item \emph{Inductive Setting}. For the inductive learning setting, a classifier is usually trained on a set of nodes and tested on others that never seen during training.
	\item \emph{Transductive Setting}. Different from inductive setting, test samples (i.e., the unlabeled nodes)  can be seen (but not their labels!) during the training procedure in transductive learning setting.
\end{itemize}
To conclude, the objective function that optimized by the classifier could be formulated as follows:
\begin{equation}
	\label{loss}
	\mathfrak{L}= \frac{1}{|\mathcal{V}_L|} \sum_{v_i \in \mathcal{V}_L}\mathcal{L}(f_{\theta} (\mathcal{G}, X), y_i)
\end{equation}
where $y_i$ is the class label of node $v_i$, and $\mathcal{L}$ could be the loss of either inductive learning or transductive learning, and the classifier $f$ with parameters $\theta$ is similarly defined with the two settings.

\nosection{\textbf{Link-level Task}}
Link-level task relates to the edge classification and link prediction. Among them, link prediction is a more challenging task and widely used in real-world, which aims to predict the connection strength of an edge, e.g., predicting the potential relationship between two specified persons in a social network, and even the new or dissolution relationship in the future. Link prediction tasks take the same input as node classification tasks, while the output is binary which indicates an edge will exist or not. Therefore the objective function of link prediction tasks could be similarly defined as Eq.(\ref{loss}) by changing $y_i \in \{0,1\}$, and replacing $\mathcal{V}_L$ with a set of labeled edges $\mathcal{E}_L$.

\nosection{\textbf{Graph-level Task}}
While treating the graph as a special form of node, the graph-level tasks are similar to node-level tasks. Take the most frequent application, graph classification, as an example, the graph $\mathcal{G}$ could be represented as $\mathcal{G}=\{ G_1, G_2,\dots, G_n \}$, where $G_i=(\mathcal{V}_i,\mathcal{E}_i)$ is a subgraph of the entire graph. The core idea to solve the graph classification problem is to learn a mapping function $\mathcal{G} \rightarrow \mathcal{C}$, here $\mathcal{C}$ represents the set of graph categories, so as to predict the class for an unseen graph more accurately. The objective function of graph classification tasks is similar with Eq.(\ref{loss}) as well, in which the labeled training set is $\mathcal{G}_L$ instead of $\mathcal{V}_L$, and $y_i$ is the category of the graph $G_i$. In real world, graph classification task plays an important role in many crucial applications, such as social and biological graph classification.

\section{Adversarial Attack}
\label{attack}
In this section, we will first introduce the definition of adversarial attack against deep learning methods on graphs. Then,  we categorize these attack models from different perspectives. Finally, we will give a clear overview on existing works.

\subsection{Definition}
\label{attackDefinition}
According to existing works on graph adversarial attacks, we summarize and give a unified formulation for them.

\nosection{Attack on Graph Data} Considering $f$ a deep learning function designed to tackle related downstream tasks. Given a set of target instances $\mathcal{T} \subseteq \mathcal{S}-\mathcal{S}_L$, where $\mathcal{S}$ could be $\mathcal{V}$, $\mathcal{E}$ or $\mathcal{G}$ respectively for different levels of tasks, and $\mathcal{S}_L$ denotes the instances with labels,\footnote{Note that attackers only focus on attacking the target instances in the test set.} the attacker aims to maximize the loss of the target node on $f$ as much as possible, resulting in the degradation of prediction performance. Generally, we can define the attack against deep learning models on graphs as:

\begin{equation}
	\label{definition}
	\begin{split}
		& \underset{\hat{G} \in \Psi(G)}{\text{maximize}} \sum_{t_i \in \mathcal{T}} \mathcal{L}(f_{\theta^{\ast}}(\hat{G}^{t_i}, X, t_i), y_i)  \\
		& s.t. \ \theta^{\ast} = \mathop{\arg \min}_{\theta} \sum_{v_j \in \mathcal{S}_L}\mathcal{L}(f_{\theta}(\tilde{G}^{v_j}, X, t_j)), y_j)
	\end{split}
\end{equation}

We denote a graph $G \in \mathcal{G}$ with node $t_i$ in it as $G^{t_i}$. $\hat{G}$ denotes the perturbed graph and $\Psi(G)$ indicates the perturbation space on $G$, and we use $\tilde{G}$ to represent original graph $G$ or a modified one $\hat{G}$, respectively.

To make the attack as imperceptible as possible, we set a metric for comparison between the graphs before and after the attack, such that:
\begin{equation}
	\begin{split}
		\mathcal{Q}(\hat{G}^{t_i}, G^{t_i}) < \epsilon \\ s.t.\ \hat{G}^{t_i} \in \Psi(G)
	\end{split}
\end{equation}
where $\mathcal{Q}$ denotes a similarity function, and $\epsilon$ is an allowed threshold of changes. Specifically, more metrics will be detailed in Section \ref{metrics}.

\subsection{Taxonomies of Attacks}
As we focus on a simple graph, different types of attacks could be conducted on the target systems. In this section, we provide taxonomies for the attack types mainly proposed by Sun et al. \cite{sun2018adversarial} and subsequently extended in our works according to various criteria. For different scenarios, we give relevant instructions and summarize them in the following:

\subsubsection{Attacker’s Knowledge}
To conduct attacks on the target system, usually an attacker will possess certain knowledge about the target models and the dataset, which helps them achieve the adversarial goal. Based on the knowledge of the target models \cite{sun2018adversarial}, we characterize different threatening levels of attacks.
\nosection{\textbf{White-box Attack}}
This is the simplest attacks while attackers possess the entire information of the target models, including the model architecture, parameters and gradient information, i.e., the target models are fully exposed to the attackers. By utilizing such rich information, attackers can easily affect the target models and bring destructive effects. However, it is impracticable in real world since it is costly to possess such complete knowledge of target models. Therefore white-box attack is less dangerous but often used to approximate the worst performance of a system under attack.
\nosection{\textbf{Gray-box Attack}}
In this case, attackers are strict to possess excessive knowledge about the target models, this reflects real world scenarios much better since it is more likely that attackers have limited access to get the information, e.g., only familiar with the architecture of the target models. Therefore, it's harder than conducting the white-box attack but more dangerous for the target models.
\nosection{\textbf{Black-box Attack}}
In contrast to white-box attack, black-box attack assumes that the attacker does not know anything about the targeted systems. Under this setting, attackers are only allowed to do black-box queries on limited samples at most. However, it will be the most dangerous attack once it works, since attackers can attack any models with limited (or no) information.

Here also comes a term \textbf{``no-box attack''} \cite{sharma2018gradient} which refers to an attack on the surrogate model based on their (limited) understanding of the target model.\footnote{We must realize that no-box is a kind of gray- or black-box attack so we don't don't have a separate category for it.} As attackers have complete knowledge of the surrogate model, the adversarial examples are generated under white-box setting. No-box attack will become a white-box attack once the attackers build a surrogate model successfully and the adversarial examples are transferable to fool other target models. Theoretically, this kind of attack can hardly affect the systems in most cases with these constraints. However, as if the surrogate model has strong transferability, the no-box attack is also destructive.

As attackers are strict with the knowledge of the target models, they also have less access to the information of the dataset $\mathcal{D}$. Based on the different levels of knowledge of the targeted system \cite{chen2017practical}, we have:
\nosection{\textbf{Perfect Knowledge}}
In this case, the dataset $\mathcal{D}$, including the entire graph structure, node features, and even the ground-truth labels of objects, are completely exposed to attackers, i.e., the attacker is assumed to know \emph{everything} about the dataset. This is impractical but the most common setting in previous studies. However, considering the damage that could be done by an attacker with perfect knowledge is critical, since it may expose the potential weaknesses of the target system in a graph.

\nosection{\textbf{Moderate Knowledge}}
The moderate knowledge case represents an attacker with less information about the dataset. This makes attackers focus more on the performance of their attacks, or even search for more information through legitimate or illegitimate methods.

\nosection{\textbf{Minimal Knowledge}}
This is the hardest one as attackers can only conduct attacks with the minimal knowledge of datasets, such as partial connections of the graph or the feature information of certain nodes. This is essential information for an attacker, otherwise it will fail even under the white-box attack setting. The minimal knowledge case represents the least sophisticated attacker and naturally with less threat level.

To conclude, we use $\mathcal{K}$ to denote the attackers' knowledge in an abstract knowledge space. Here $\mathcal{K}$ is consisting of three main parts, the dataset knowledge, the training algorithm and the learned parameters of target models. For instance, $\mathcal{K}=(\mathcal{D},f,\theta)$ denotes white-box attack with perfect knowledge, and $\mathcal{K}=(\mathcal{D}_{train},\hat{f},\hat{\theta})$ denotes gray- or black-box attack and with moderate knowledge, where $\hat{f}$ and $\hat{\theta}$ come from the surrogate model and $\mathcal{D}_{train}\subset \mathcal{D}$ denotes the training dataset.

\begin{figure}
	\centering
	\includegraphics[width=70mm]{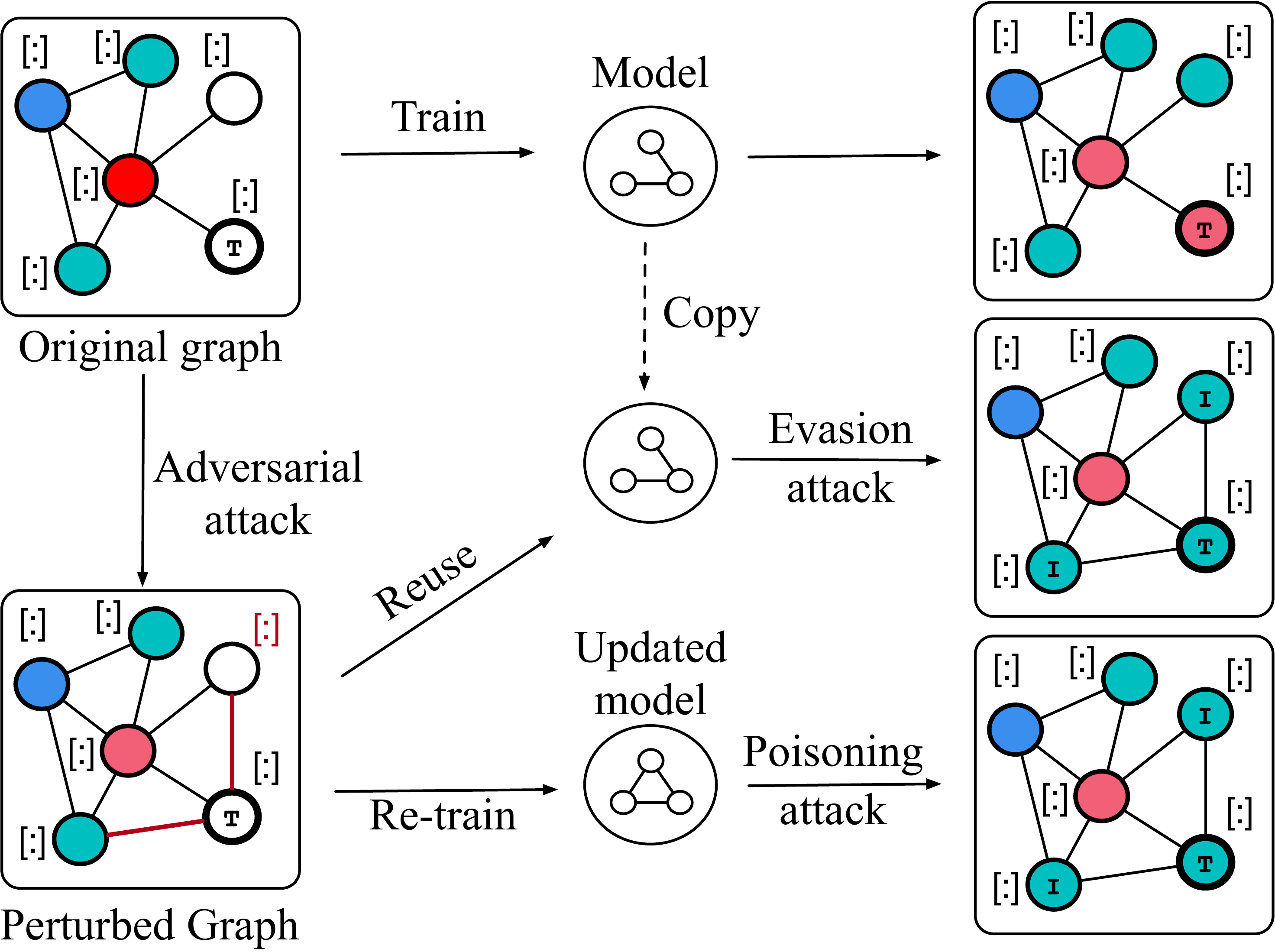}
	\caption{An example of the evasion attack and the poisoning attack. (Image Credit: Z{\"u}nger et al. \protect\cite{zugner2018adversarial}) }
	\label{Fig2}
\end{figure}

\subsubsection{Attacker's Goal}
The goal generally distinguishes three different types of attacks, i.e., \emph{security violation}, \emph{attack specificity} and \emph{error specificity} \cite{biggio2018wild}.	They are not mutually exclusive in fact, and more details are discussed below.

\nosection{\textbf{Security Violation}}
Security violation can be categorized into \emph{availability attack}, \emph{integrity attack} and \emph{others}. For availability attack, the the attacker attempts to destroy the function of the system, thereby impairing its normal operation. The damage is global, that is, it attacks the overall performance of the whole system. For integrity attack, the attackers' purpose is to bypass or fool the detection of the system, which is different from availability attack in that it does not destroy the normal operation of the system. There are other goals, such as reverse-engineering model information to gain privacy knowledge.

\nosection{\textbf{Error Specificity}}
Take node classification task as an example, the error specific attack aims to misclassify the predictions as specific labels, while unspecific attack does not care what the prediction is, the attack is considered successful as long as the prediction is wrong.

\nosection{\textbf{Attack Specificity}}
This perspective focuses on the range of the attack, which can divide attacks into \emph{targeted attack} and \emph{non-targeted attack (general attack)}. The targeted attack focuses on a specific subset of nodes (usually a target node), while the non-targeted attack is undifferentiated and global. With reference to Eq.(\ref{definition}), the difference is the domain of $\mathcal{T} \subset \mathcal{S} - \mathcal{S}_L$ or $\mathcal{T} = \mathcal{S} - \mathcal{S}_L$.

It is worth noting that in some other fields (e.g., computer vision), \emph{targeted attack} refers to the specific error attack, and \emph{non-targeted attack} the unspecific error attack. In the field of graph adversarial learning, we suggest that distinguishing the targeted attack based on the attack range, and whether it is error specific based on the consequence.

\subsubsection{Attacker's Capability}
\label{period}
Attacks can be divided into poisoning attack and evasion attack according to adversaries' capabilities, which are occurred at different stages of the attacks.
\nosection{\textbf{Poisoning Attack}}
Poisoning attacks (a.k.a training-time attacks) try to affect the performance of the target models by modifying the dataset in the training stage, i.e., the target models are trained on the poisoned datasets. Since transductive learning is widely used in most graph analysis tasks, the test samples (but not their labels) are participated in the training stage, which leads to the popularity of poisoning attacks. Under this scenario, the parameters of target models are retrained after the training dataset being modified, thus we can define poisoning attacks according to Eq.(\ref{definition}):
\begin{equation}
	\label{poisoning}
	\begin{aligned}
		 & \underset{\hat{G} \in \Psi(G)} {\text{maximize}} \sum_{t_i \in \mathcal{T}} \mathcal{L}(f_{\theta^*}(\hat{G}^{t_i},X,t_i),y_i)   \\
		 & s.t. \ \theta^{\ast} = \mathop{\arg \min}_{\theta} \sum_{v_j \in \mathcal{S}_L}\mathcal{L}(f_{\theta}(\hat{G}^{v_j},X,v_j), y_j)
	\end{aligned}
\end{equation}

\nosection{\textbf{Evasion Attack}}
While poisoning attacks focus on the training phase, evasion attacks (a.k.a test-time attacks) tend to add adversarial examples in test time. Evasion attacks occur after the target model is well trained on a clean graph, i.e., the learned parameters are fixed during evasion attacks. Therefore, we can define the formulation of evasion attacks by changing part of Eq.(\ref{poisoning}) slightly:
\begin{equation}
	\begin{aligned}
		\theta^{\ast} = \mathop{\arg \min}_{\theta} \sum_{v_j \in \mathcal{S}_L}\mathcal{L}(f_{\theta}(G^{v_j},X,v_j), y_j)
	\end{aligned}
\end{equation}

In Figure \ref{Fig2}, we show a toy example of the poisoning attack and the evasion attack. In most cases the poisoning attack does not work well because the model is retrained to alleviate the adversarial impacts.

\subsubsection{Attack Strategy}
For attacking a target model on graph data, attackers may have a line of strategies to achieve their adversarial goals. In most instances, they will focus on the graph structure or node/edge features. Based on the strategy applied on a graph, we have:
\nosection{\textbf{Topology Attack}}
Attackers are mainly focus on the the topology of the graph, a.k.a, structure attack. For example, they are allowed to add or remove some edges legally between nodes in the graph to fool the target system. To specify this, we define an attack budget $\Delta \in \mathbb{N}$, thus the definition of the topology attack is:
\begin{equation}
	\begin{aligned}
		 & \underset{\hat{G} \in \Psi(G)} {\text{maximize}} \sum_{t_i \in \mathcal{T}} \mathcal{L}(f_{\theta^*}(\hat{G}^{t_i},X,t_i),y_i) \\
		 & s.t. \  \sum_{u<v} |A_{u,v}-A'_{u,v} | \leq \Delta
	\end{aligned}
\end{equation}
where $A'$ is the adjacency matrix of the perturbed graph, and $\theta^*$ is discussed in Section \ref{period}.

\nosection{\textbf{Feature Attack}}
Although the topology attack is more common, attackers can conduct feature attacks as well. Under this setting, the features of specified objects will be changed. However, unlike graph structure, the node/edge features could be either binary or continuous, i.e., $X \in \{0,1\}^{N \times F_{(\cdot)}}$ or $X \in \mathbb{R}^{N \times F_{(\cdot)}}$. For binary features, attackers can flip them like edges, while for continuous features, attackers can add a small value on them, respectively. Thus the definition of feature attacks is:
\begin{equation}
	\begin{aligned}
		 & \underset{\hat{G} \in \Psi(G)} {\text{maximize}} \sum_{t_i \in \mathcal{T}} \mathcal{L}(f_{\theta^*}(\hat{G}^{t_i},X,t_i),y_i) \\
		 & s.t. \  \sum_u \sum_j  |X_{u,j}-X'_{u,j} | \leq \Delta
	\end{aligned}
\end{equation}
where $X'$ is similarly defined, and  here $\Delta \in \mathbb{N}$ for binary features and $\Delta \in \mathbb{R}$ for continuous features.

\nosection{\textbf{Hybrid}}
Usually, attackers will conduct both attack strategies at the same time to exert more powerful impact. Besides, they could even add several fake nodes (with fake labels), which will have their own features and relationship with other benign instances. For example, some fake users will be added into the recommendation system and affect the results of the system \cite{chen2019data,hou2019alphacyber}. Therefore, we conclude a unified formulation as follows:
\begin{equation}
	\begin{aligned}
		 & \underset{\hat{G} \in \Psi(G)} {\text{maximize}} \sum_{t_i \in \mathcal{T}} \mathcal{L}(f_{\theta^*}(\hat{G}^{t_i},X,t_i),y_i) \\
		 & s.t. \  \sum_{u<v} |A_{u,v}-A'_{u,v} | + \sum_u \sum_j  |X_{u,j}-X'_{u,j} |\leq \Delta
	\end{aligned}
\end{equation}
$A'$ and $X'$ may not have the same dimension with $A$ and $X$ respectively if some fake nodes are added into the graph.

\subsubsection{Attacker’s Manipulation}
Although attackers may have enough knowledge about the target system, they may not always have access to manipulation all of the dataset. Besides, different manipulations may have different budget cost. For example, in an e-commerce system \cite{DBLP:conf/ijcai/ChenL0GZ19}, attackers could only purchase more items (add edges) but unable to delete the purchase records (remove edges). Based on different manipulation, we have:
\nosection{\textbf{Add}}
In a graph, attackers could add edges between different nodes, or add features on specified nodes/edges. This is the most simplest manipulation and naturally has lowest budget in most scenarios.
\nosection{\textbf{Remove}}
In a graph, attackers could remove edges between different nodes, or remove features on specified nodes/edges. This kind of manipulation will be harder than ``add'', since attackers may not have enough access to execute the ``remove'' manipulation.
\nosection{\textbf{Rewiring}}
The manipulations mentioned above may become more noticeable for the target system if $\Delta$ is larger. To address this problem, the rewiring operation is proposed in a less noticeable way than simple adding/removing manipulation \cite{ma2019attacking, waniek2018attack}. For instance, attackers can add an edge connected with a target node, while remove an edge connected with it at the same time. Each time of rewiring manipulation related to two single operations, which can preserve some basic graph properties (e.g., degree) and naturally more unnoticeable.

To conclude, we define an operation set $\mathcal{O}=\{\mathcal{O}_{Add},\mathcal{O}_{Rem},\mathcal{O}_{Rew}\}$, representing the above three manipulations of attackers, respectively.

\subsubsection{Attack Algorithm}
Generally speaking, current methods of adversarial attacks on generating adversarial examples are mainly based on the gradient information, either from the target model (white-box attack) or the surrogate model (black- or gray-box attack). Beyond that, there are several methods generating adversarial examples based on other algorithms. From the perspective of attack algorithm, we have:
\nosection{\textbf{Gradient-based Algorithm}}
Intuitively, gradient-based algorithm is simple but effective. The core idea is that: fix the parameters of a trained model, and regard the input as a hyperparameter to optimize. Similar to the training process, attackers could use the partial derivative of loss $\mathcal{L}$ with respect to edges (topology attack) or features (feature attack), to decide how to manipulate the dataset. However, gradients could not applied directly into the input data due to the discreteness of the graph data, instead, attackers often choose the one with greatest absolute gradients, and manipulate it with a proper value. While most deep learning models are optimized by gradients, on the contrary, attackers could destroy them by gradients as well.

\nosection{\textbf{Non-gradient-based Algorithm}}
In addition to gradient information, attackers could generate adversarial examples in other ways. For example, from the perspective of the genetic algorithm, attackers can choose the population (adversarial examples) with highest fitness score (e.g., erroneous outputs of the target/surrogate model) generations by generations. Besides, reinforcement learning algorithms are also commonly used to solve this issue. Reinforcement learning based attack methods \cite{dai2018adversarial,ma2019attacking} will learn the generalizable adversarial examples within an action space. Moreover, adversarial examples can even generated by a well-designed generative model.

\subsubsection{Target Task}
As discussed in Section \ref{task}, there exists three major tasks in graph domains. According to different levels of tasks, we can divide existing attack methods into the following three categories, respectively.

\nosection{\textbf{Node-relevant Task}}
Currently, there are several attack models against node-relevant tasks \cite{zugner2018adversarial,bojchevski2019adversarial,xu2019topology,zugner2019adversarial,wang2018attack,chen2018fast,dai2018adversarial,dai2018adversarial,xuan2019unsupervised,bose2019generalizable,wang2019attacking,waniek2018attack,chen2018link,sun2018data,zhou2019attacking,wu2019adversarial}. Typically, Bojchevski et al. and Hou et al. \cite{bojchevski2019adversarial,hou2019alphacyber} utilize random walk \cite{perozzi2014deepwalk} as an surrogate model to attack node embedding,  Dai et al. \cite{dai2018adversarial} uses a reinforcement learning based framework to disturb node classification tasks. In general, most of the existing works address in node classification task due to its ubiquity in real world.

\nosection{\textbf{Link-relevant Task}}
Many relationships in real world can be represented by graph. For some of these graphs, such as social graph, are always dynamic in reality. Therefore, the link prediction on the graph comes into being, in order to predict the change of the edge. Link prediction is also the most common application of link-relevant tasks and there are many attack related studies have been proposed \cite{chen2018link,waniek2018attack,zhou2019attacking,sun2018data,zhou2019attacking,chen2019data,christakopoulou2018adversarial,chen2019time}.

\nosection{\textbf{Graph-relevant Task}}
Graph-relevant tasks treat graph as a basic unit. Compared to node- or link-relevant tasks, it is macro and large-scale. The application of graph-relevant methods are more inclined to the research community of biology, chemistry, environment, materials, etc. In this task, the model tries to extract features of nodes and spatial structures to represent a graph, so as to achieve downstream operations such as graph classification or clustering. Similar with  Eq.(\ref{definition}), the targets of attack should be an graphs, while $y$ is determined by the specific task. For graph-relevant tasks, some attack researches have also appeared \cite{ma2019attacking,chen2019ga,chen2017practical}.

\begin{table*}[]
	\centering
	\caption{Related works on attack in details.}
	\label{attack_table}
	\scalebox{0.75}{
		\begin{tabular}{c|cccccccccc}
			\hline
			\hline
			Reference                             &
			Model                                 &
			Algorithm                             &
			Target Task                           &
			Target Model                          &
			Baseline                              &
			Metric                                &
			Dataset                               & \\

			\hline
			\cite{chang2019restricted}            &
			\begin{tabular}[c]{@{}c@{}} GF-Attack \end{tabular}            &
			\begin{tabular}[c]{@{}c@{}} Graph signal processing\end{tabular}            &
			\begin{tabular}[c]{@{}c@{}} Node classification \end{tabular}            &
			\begin{tabular}[c]{@{}c@{}} GCN, SGC,\\DeepWalk, LINE \end{tabular}            &
			\begin{tabular}[c]{@{}c@{}} Random, \\Degree, \\RL-S2V,\\ $A_{class}$ \end{tabular}            &
			\begin{tabular}[c]{@{}c@{}} Accuracy \end{tabular}            &
			\begin{tabular}[c]{@{}c@{}} Cora,\\Citeseer,\\Pubmed \end{tabular}              \\

			\hline
			\cite{wu2019adversarial}              &
			\begin{tabular}[c]{@{}c@{}} IG-FGSM,\\IG-JSMA \end{tabular}            &
			\begin{tabular}[c]{@{}c@{}} Gradient-based GCN\end{tabular}            &
			\begin{tabular}[c]{@{}c@{}} Node classification \end{tabular}            &
			\begin{tabular}[c]{@{}c@{}} GCN \end{tabular}            &
			\begin{tabular}[c]{@{}c@{}} FGSM,\\ JSMA,\\ Nettack \end{tabular}            &
			\begin{tabular}[c]{@{}c@{}} Classification margin,\\Accuracy \end{tabular}            &
			\begin{tabular}[c]{@{}c@{}} Cora,\\Citeseer,\\PolBlogs \end{tabular}              \\

			\hline
			\cite{chen2019multiscale}             &
			\begin{tabular}[c]{@{}c@{}} EPA \end{tabular}            &
			\begin{tabular}[c]{@{}c@{}} Genetic algorithm \end{tabular}            &
			\begin{tabular}[c]{@{}c@{}} Community detection \end{tabular}            &
			\begin{tabular}[c]{@{}c@{}} GRE, INF,\\ LOU \end{tabular}            &
			\begin{tabular}[c]{@{}c@{}} $A_Q$, $A_S$, $A_B$,\\$A_D$, $D_S$, $D_W$ \end{tabular}            &
			\begin{tabular}[c]{@{}c@{}} NMI,\\ ARI \end{tabular}            &
			\begin{tabular}[c]{@{}c@{}} Synthetic networks,\\Football, Email, \\Polblogs \end{tabular}              \\

			\hline
			\cite{zugner2019adversarial}          &
			\begin{tabular}[c]{@{}c@{}} Meta-Self, \\Meta-Train \end{tabular}            &
			\begin{tabular}[c]{@{}c@{}} Gradient-Based GCN \end{tabular}            &
			\begin{tabular}[c]{@{}c@{}} Node  classification \end{tabular}            &
			\begin{tabular}[c]{@{}c@{}} GCN,\\CLN,\\Deepwalk \end{tabular}            &
			\begin{tabular}[c]{@{}c@{}} DICE,\\Nettack,\\First-order \end{tabular}           &
			\begin{tabular}[c]{@{}c@{}} Misclassification rate,\\Accuracy \end{tabular}           &
			\begin{tabular}[c]{@{}c@{}} Cora-ML,\\Citeseer,\\PolBlogs,\\PubMed, \end{tabular}             \\

			\hline
			\cite{bojchevski2019adversarial}      &
			\begin{tabular}[c]{@{}c@{}} $\mathcal{A}_{DW2}$,\\$\mathcal{A}_{DW3}$ \end{tabular}           &
			\begin{tabular}[c]{@{}c@{}} Gradient-based random walk \end{tabular}           &
			\begin{tabular}[c]{@{}c@{}}Node classification,\\Link prediction\end{tabular}           &
			\begin{tabular}[c]{@{}c@{}} Deepwalk \end{tabular}           &
			\begin{tabular}[c]{@{}c@{}}$\mathcal{B}_{rnd}$\\$\mathcal{B}_{eig}$\\$\mathcal{B}_{deg}$ \end{tabular}           &
			\begin{tabular}[c]{@{}c@{}} F1 score,\\Classification margin \end{tabular}           &
			\begin{tabular}[c]{@{}c@{}}Cora,\\Citeseer,\\PolBlogs\end{tabular}             \\

			\hline
			\cite{chen2019time}                   &
			\begin{tabular}[c]{@{}c@{}} TGA-Tra,\\ TGA-Gre \end{tabular}           &
			\begin{tabular}[c]{@{}c@{}} Gradient-based DDNE \end{tabular}           &
			\begin{tabular}[c]{@{}c@{}}Link prediction\end{tabular}           &
			\begin{tabular}[c]{@{}c@{}} DDNE,\\ctRBM,\\GTRBM,\\dynAERNN \end{tabular}           &
			\begin{tabular}[c]{@{}c@{}}Random, \\DGA,\\CNA \end{tabular}           &
			\begin{tabular}[c]{@{}c@{}} ASR, AML \end{tabular}           &
			\begin{tabular}[c]{@{}c@{}}RADOSLAW,\\LKML,\\FB-WOSN\end{tabular}             \\

			\hline
			\cite{ma2019attacking}                &
			\begin{tabular}[c]{@{}c@{}} ReWatt \end{tabular}           &
			\begin{tabular}[c]{@{}c@{}} Reinforcement learning\\based on GCN \end{tabular}           &
			\begin{tabular}[c]{@{}c@{}} Graph\\classification \end{tabular}           &
			\begin{tabular}[c]{@{}c@{}} GCN \end{tabular}           &
			\begin{tabular}[c]{@{}c@{}} RL-S2V,\\Random \end{tabular}           &
			\begin{tabular}[c]{@{}c@{}} ASR \end{tabular}           &
			\begin{tabular}[c]{@{}c@{}} REDDIT-\\MULTI-12K,\\REDDIT-MULTI-5K,\\IMDB-MULTI \end{tabular}             \\

			\hline
			\cite{xu2019topology}                 &
			\begin{tabular}[c]{@{}c@{}} PGD,\\Min-Max \end{tabular}           &
			\begin{tabular}[c]{@{}c@{}} Gradient-based GCN \end{tabular}           &
			\begin{tabular}[c]{@{}c@{}} Node classification \end{tabular}           &
			\begin{tabular}[c]{@{}c@{}} GCN \end{tabular}           &
			\begin{tabular}[c]{@{}c@{}} DICE,\\Meta-Self,\\Greedy \end{tabular}           &
			\begin{tabular}[c]{@{}c@{}} Misclassification\\rate \end{tabular}           &
			\begin{tabular}[c]{@{}c@{}} Cora,\\Citeseer \end{tabular}             \\

			\hline
			\cite{xuan2019unsupervised}           &
			\begin{tabular}[c]{@{}c@{}} EDA \end{tabular}           &
			\begin{tabular}[c]{@{}c@{}} Genetic algorithm\\based on Deepwalk \end{tabular}           &
			\begin{tabular}[c]{@{}c@{}} Node classification,\\Community detection \end{tabular}           &
			\begin{tabular}[c]{@{}c@{}} HOPE,\\LPA,\\EM,\\Deepwalk \end{tabular}           &
			\begin{tabular}[c]{@{}c@{}} Random,\\DICE,\\RLS,\\DBA \end{tabular}           &
			\begin{tabular}[c]{@{}c@{}} NMI,\\Micro-F1,\\Macro-F1\end{tabular}           &
			\begin{tabular}[c]{@{}c@{}} Karate,\\Game,\\Dolphin \end{tabular}             \\

			\hline
			\cite{bose2019generalizable}          &
			\begin{tabular}[c]{@{}c@{}} DAGAER \end{tabular}           &
			\begin{tabular}[c]{@{}c@{}} Generative model based\\on VGAE \end{tabular}           &
			\begin{tabular}[c]{@{}c@{}} Node classification \end{tabular}           &
			\begin{tabular}[c]{@{}c@{}} GCN \end{tabular}           &
			\begin{tabular}[c]{@{}c@{}} Nettack \end{tabular}           &
			\begin{tabular}[c]{@{}c@{}} ASR \end{tabular}           &
			\begin{tabular}[c]{@{}c@{}} Cora,\\Citeseer\end{tabular}             \\

			\hline
			\cite{zhang2019data}                  &
			\begin{tabular}[c]{@{}c@{}} - \end{tabular}           &
			\begin{tabular}[c]{@{}c@{}} Knowledge embedding \end{tabular}           &
			\begin{tabular}[c]{@{}c@{}} Fact plausibility\\prediction \end{tabular}           &
			\begin{tabular}[c]{@{}c@{}} TransE,\\TransR,\\RESCAL \end{tabular}           &
			\begin{tabular}[c]{@{}c@{}} Random \end{tabular}           &
			\begin{tabular}[c]{@{}c@{}} MRR, HR@K \end{tabular}           &
			\begin{tabular}[c]{@{}c@{}} FB15k,\\WN18 \end{tabular}             \\

			\hline
			\cite{wang2019attacking}              &
			\begin{tabular}[c]{@{}c@{}} - \end{tabular}           &
			\begin{tabular}[c]{@{}c@{}} Based on LinLBP \end{tabular}           &
			\begin{tabular}[c]{@{}c@{}} Node classification,\\Evasion \end{tabular}           &
			\begin{tabular}[c]{@{}c@{}} LinLBP, JW, GCN\\LBP, RW, LINE\\DeepWalk\\Node2vec \end{tabular}           &
			\begin{tabular}[c]{@{}c@{}} Random,\\Nettack \end{tabular}           &
			\begin{tabular}[c]{@{}c@{}} FNR,\\FPR \end{tabular}           &
			\begin{tabular}[c]{@{}c@{}} Facebook, Enron,\\Epinions,\\Twitter,\\Google+ \end{tabular}             \\

			\hline
			\cite{chen2019ga}                     &
			\begin{tabular}[c]{@{}c@{}} Q-Attack \end{tabular}           &
			\begin{tabular}[c]{@{}c@{}} Genetic algorithm \end{tabular}           &
			\begin{tabular}[c]{@{}c@{}} Community detection \end{tabular}           &
			\begin{tabular}[c]{@{}c@{}} FN, Lou, SOA,\\LPA, INF,\\Node2vec+KM \end{tabular}           &
			\begin{tabular}[c]{@{}c@{}} Random,\\CDA,\\DBA \end{tabular}           &
			\begin{tabular}[c]{@{}c@{}} NMI,\\Modularity Q \end{tabular}           &
			\begin{tabular}[c]{@{}c@{}} Karate,\\Dolphins,\\Football,\\Polbooks,\end{tabular}             \\

			\hline
			\cite{hou2019alphacyber}              &
			\begin{tabular}[c]{@{}c@{}} HG-attack \end{tabular}           &
			\begin{tabular}[c]{@{}c@{}} Label propagation algorithm, \\  Nodes injection \end{tabular}           &
			\begin{tabular}[c]{@{}c@{}} Malware detection \end{tabular}           &
			\begin{tabular}[c]{@{}c@{}} Orig-HGC \end{tabular}           &
			\begin{tabular}[c]{@{}c@{}} AN-Attack \end{tabular}           &
			\begin{tabular}[c]{@{}c@{}} TP, TN, FP, FN, F1, \\Precision, Recall, Accuracy\end{tabular}           &
			\begin{tabular}[c]{@{}c@{}} Tencent Security \\Lab Dataset \end{tabular}             \\

			\hline
			\cite{chen2019data}                   &
			\begin{tabular}[c]{@{}c@{}} UNAttack \end{tabular}           &
			\begin{tabular}[c]{@{}c@{}} Gradient-based similarity method, \\  Nodes injection \end{tabular}           &
			\begin{tabular}[c]{@{}c@{}} Recommendation \end{tabular}           &
			\begin{tabular}[c]{@{}c@{}} Memory-based CF,\\BPRMF, NCF \end{tabular}           &
			\begin{tabular}[c]{@{}c@{}} Random,\\ Average,\\ Popular,\\ Co-visitation \end{tabular}           &
			\begin{tabular}[c]{@{}c@{}} Hit@K\end{tabular}           &
			\begin{tabular}[c]{@{}c@{}} Filmtrust,\\Movielens,\\Amazon \end{tabular}             \\

			\hline
			\cite{christakopoulou2018adversarial} &
			\begin{tabular}[c]{@{}c@{}} - \end{tabular}           &
			\begin{tabular}[c]{@{}c@{}} Gradient-based GAN and MF, \\  Nodes injection \end{tabular}           &
			\begin{tabular}[c]{@{}c@{}} Recommendation \end{tabular}           &
			\begin{tabular}[c]{@{}c@{}} MF \end{tabular}           &
			\begin{tabular}[c]{@{}c@{}} - \end{tabular}           &
			\begin{tabular}[c]{@{}c@{}} Attack difference,\\TVD, JS, Est., \\Rank loss @K,\\Adversarial loss \end{tabular}           &
			\begin{tabular}[c]{@{}c@{}} MovieLens 100k,\\MovieLens 1M \end{tabular}             \\

			\hline
			\cite{wang2018attack}                 &
			\begin{tabular}[c]{@{}c@{}} Greedy GAN \end{tabular}           &
			\begin{tabular}[c]{@{}c@{}} Gradient-based GCN and GAN \end{tabular}           &
			\begin{tabular}[c]{@{}c@{}} Node\\classification \end{tabular}           &
			\begin{tabular}[c]{@{}c@{}} GCN \end{tabular}           &
			\begin{tabular}[c]{@{}c@{}} Random \end{tabular}           &
			\begin{tabular}[c]{@{}c@{}} Accuracy,\\F1 score, ASR \end{tabular}           &
			\begin{tabular}[c]{@{}c@{}} Cora,\\Citeseer\end{tabular}             \\

			\hline
			\cite{waniek2018attack}               &
			\begin{tabular}[c]{@{}c@{}} CTR, OTC \end{tabular}           &
			\begin{tabular}[c]{@{}c@{}} Neighbor score based\\on graph structure\end{tabular}           &
			\begin{tabular}[c]{@{}c@{}} Link prediction \end{tabular}           &
			\begin{tabular}[c]{@{}c@{}} Traditional\\link prediction\\algorithms\end{tabular}           &
			\begin{tabular}[c]{@{}c@{}} - \end{tabular}           &
			\begin{tabular}[c]{@{}c@{}} AUC, AP \end{tabular}           &
			\begin{tabular}[c]{@{}c@{}} WTC 9/11,\\ScaleFree,\\Facebook,\\Random network\end{tabular}             \\

			\hline
			\cite{chen2018link}                   &
			\begin{tabular}[c]{@{}c@{}} IGA \end{tabular}           &
			\begin{tabular}[c]{@{}c@{}} Gradient-based GAE\end{tabular}           &
			\begin{tabular}[c]{@{}c@{}} Link prediction \end{tabular}           &
			\begin{tabular}[c]{@{}c@{}} GAE,LRW\\DeepWalk,\\Node2vec,\\CN, Random, Katz \end{tabular}           &
			\begin{tabular}[c]{@{}c@{}} RAN,\\DICE,\\GA \end{tabular}           &
			\begin{tabular}[c]{@{}c@{}} ASR,\\ AML \end{tabular}           &
			\begin{tabular}[c]{@{}c@{}} NS,\\Yeast,\\FaceBook \end{tabular}             \\

			\hline
			\cite{dai2018adversarial}             &
			\begin{tabular}[c]{@{}c@{}} RL-S2V \end{tabular}           &
			\begin{tabular}[c]{@{}c@{}} Reinforcement learning \end{tabular}           &
			\begin{tabular}[c]{@{}c@{}} Node/Graph\\ Classification \end{tabular}           &
			\begin{tabular}[c]{@{}c@{}} GCN,\\GNN \end{tabular}           &
			\begin{tabular}[c]{@{}c@{}} Random\\sampling \end{tabular}           &
			\begin{tabular}[c]{@{}c@{}} Accuracy \end{tabular}           &
			\begin{tabular}[c]{@{}c@{}} Citeseer, Cora,\\Finance,\\Pubmed  \end{tabular}             \\

			\hline
			\cite{zugner2018adversarial}          &
			\begin{tabular}[c]{@{}c@{}} Nettack \end{tabular}           &
			\begin{tabular}[c]{@{}c@{}} Greedy search \& gradient \\based on GCN \end{tabular}           &
			\begin{tabular}[c]{@{}c@{}} Node classification \end{tabular}           &
			\begin{tabular}[c]{@{}c@{}} GCN,\\CLN,\\Deepwalk \end{tabular}           &
			\begin{tabular}[c]{@{}c@{}} Rnd,\\FGSM \end{tabular}           &
			\begin{tabular}[c]{@{}c@{}} Classification margin,\\Accuracy, \end{tabular}           &
			\begin{tabular}[c]{@{}c@{}} Cora-ML,\\Citeseer,\\PolBlogs \end{tabular}             \\

			\hline
			\cite{chen2018fast}                   &
			\begin{tabular}[c]{@{}c@{}} FGA \end{tabular}           &
			\begin{tabular}[c]{@{}c@{}} Gradient-based GCN \end{tabular}           &
			\begin{tabular}[c]{@{}c@{}} Node classification,\\Community detection \end{tabular}           &
			\begin{tabular}[c]{@{}c@{}} GCN, GraRep,\\DeepWalk,\\Node2vec,\\LINE, GraphGAN \end{tabular}           &
			\begin{tabular}[c]{@{}c@{}} Random,\\DICE,\\Nettack \end{tabular}           &
			\begin{tabular}[c]{@{}c@{}} ASR, AML \end{tabular}           &
			\begin{tabular}[c]{@{}c@{}} Cora,\\Citeseer,\\PolBlogs, \end{tabular}             \\

			\hline
			\cite{sun2018data}                    &
			\begin{tabular}[c]{@{}c@{}} Opt-attack \end{tabular}           &
			\begin{tabular}[c]{@{}c@{}} Gradient-based Deepwalk and \end{tabular}           &
			\begin{tabular}[c]{@{}c@{}} Link prediction \end{tabular}           &
			\begin{tabular}[c]{@{}c@{}} DeepWalk,\\LINE, SC\\Node2vec, GAE \end{tabular}           &
			\begin{tabular}[c]{@{}c@{}} Random,\\ PageRank,\\Degree sum,\\Shortest path \end{tabular}           &
			\begin{tabular}[c]{@{}c@{}} AP, \\Similarity Score\end{tabular}           &
			\begin{tabular}[c]{@{}c@{}} Facebook,\\Cora,\\Citeseer \end{tabular}             \\

			\hline
			\cite{zhou2019attacking}              &
			\begin{tabular}[c]{@{}c@{}} Approx-Local \end{tabular}           &
			\begin{tabular}[c]{@{}c@{}} Similarity methods \end{tabular}           &
			\begin{tabular}[c]{@{}c@{}} Link prediction \end{tabular}           &
			\begin{tabular}[c]{@{}c@{}} Local\& global\\similarity\\metrics\end{tabular}           &
			\begin{tabular}[c]{@{}c@{}} RandomDel,\\GreedyBase \end{tabular}           &
			\begin{tabular}[c]{@{}c@{}} Katz similarity,\\ACT distance,\\Similarity score \end{tabular}           &
			\begin{tabular}[c]{@{}c@{}} Random network,\\Facebook \end{tabular}             \\

			\hline
			\cite{chen2017practical}              &
			\begin{tabular}[c]{@{}c@{}}Targeted noise injection,\\Small community attack \end{tabular}           &
			\begin{tabular}[c]{@{}c@{}} Noise injection \end{tabular}           &
			\begin{tabular}[c]{@{}c@{}} Graph clustering,\\Community detection \end{tabular}           &
			\begin{tabular}[c]{@{}c@{}} SVD, Node2vec,\\Community\\detection algorithms \end{tabular}           &
			\begin{tabular}[c]{@{}c@{}} - \end{tabular}           &
			\begin{tabular}[c]{@{}c@{}} ASR, FPR\end{tabular}           &
			\begin{tabular}[c]{@{}c@{}} Reverse,\\Engineered,\\DGA,\\Domains,\\NXDOMAIN \end{tabular}             \\

			\hline
			\hline
		\end{tabular}
	}
\end{table*}

\subsection{Summary: Attack on Graph}
In this part, we will discuss the main contributions of current works, and point out the limitations to be overcome, and propose several open questions in this area. Specifically, we cover these released papers and its characteristic in Table \ref{attack_table}.

\subsubsection{Major Contributions}
So far, most of the existing works in the attack scenario are mainly based on the gradients,  either the adjacency matrix or feature matrix, leading to topology attack and feature attack, respectively. However, the gradients' information of the target model is hard to acquire, instead, attackers will train a surrogate model to extract the gradients. In addition to gradient-based algorithms, several heuristic methods are proposed to achieve the goal of attack, such as  genetic algorithm \cite{whitley1994genetic} and reinforcement learning \cite{sutton2018reinforcement} based algorithms. According to different tasks in graph analysis, we summarize the major contributions of current works in the following.

\nosection{\textbf{Node-relevant Task}}
Most of the researches focus on the node classification task. The work \cite{zugner2018adversarial} is the first to study the adversarial attack on graph data, using an efficient greedy search method to perform perturbations on node features and graph structure and attacking traditional graph learning model. From the perspective of gradients, there are several works \cite{xu2019topology,zugner2019adversarial,wang2018attack,chen2018fast,dai2018adversarial,wu2019adversarial} focus on the topology attack, adding/removing edges between nodes based on the gradients information form various surrogate models. Specially, Xu et al. \cite{xu2019topology} present a novel optimized-based attack method that uses the gradients of surrogate model and facilitates the difficulty of tackling discrete graph data; Z\"ugner et al. \cite{zugner2019adversarial} use meta-gradients to solve the bilevel problem of poisoning a graph; Wang et al. \cite{wang2018attack} propose a greedy method based on Generative Adversarial Network (GAN) \cite{goodfellow2014generative} to generate adjacency and feature matrices of fake nodes, which will be injected to a graph to misclassify the target models; Chen et al. and Dai et al. \cite{chen2018fast, dai2018adversarial} both use GCN as a surrogate model to extract gradients information and thus generating an adversarial graph; Wu et al. \cite{wu2019adversarial} argue that integrated gradients can better reflect the effect of perturbing certain features or edges. Moreover, Dai et al. \cite{dai2018adversarial} consider evasion attacks on the task of node classification and graph classification, and proposes two effective attack methods based on the reinforcement learning and genetic algorithms, respectively. Taking Deepwalk \cite{perozzi2014deepwalk} as a base method,  Bojchevski et al. \cite{bojchevski2019adversarial} and Xuan et al. \cite{xuan2019unsupervised} propose a eigen decomposition and genetic algorithm based strategy to attack the network embedding, respectively. Also, Bose et al. \cite{bose2019generalizable} design a unified encoder-decoder framework from the generative perspective, which can be employed to attack diverse domains (images, text and graphs), Wang et al. \cite{wang2019attacking} propose a threat model to manipulate the graph structure to evade detection by solving a graph-based optimization problem efficiently. Considering real-world application scenarios, Hou et al. \cite{hou2019alphacyber} propose an algorithm that allows malware to evade detection by injecting nodes (apps).

\nosection{\textbf{Link-relevant Task}}
Link prediction is another fundamental research problems in network analysis. In this scenario,  Waniek et al. \cite{waniek2018attack} study the link connections, and propose heuristic algorithms to evade the detection by rewiring operations. Furthermore, Chen et al. \cite{chen2018link} put forward a novel iterative gradient attack method based on a graph auto-encoder framework. Similarly, Chen et al. \cite{chen2019time} also exploit the gradients information of surrogate model, and firstly study the works about adversarial attacks on dynamic network link prediction (DNLP). Besides, Sun et al. \cite{sun2018data} focus on poisoning attacks and propose a unified optimization framework based on projected gradient descent. In the recommendation scenario, considering the interactions between users and items as a graph and treat it as a link prediction task, Christakopoulou et al. \cite{christakopoulou2018adversarial} and Chen et al. \cite{chen2019data} propose the method of injecting fake users to degrade the recommendation performance of the system.

\nosection{\textbf{Graph-relevant Task}}
Few works study the adversarial attacks on this scenario, Ma et al. \cite{ma2019attacking} propose a rewiring operation based algorithm, which uses reinforcement learning to learn the attack strategy on the task of graph classification. Besides, Chen et al. \cite{chen2019ga} introduce the problem of community detection and proposes a genetic algorithm based method. Chen et al. \cite{chen2017practical} focus on graph clustering and community detection scenarios, devise generic attack algorithms based on noise injection and demonstrates their effectiveness against a real-world system.

\subsubsection{Current Limitations}
\label{attack_limitation}
Despite the remarkable achievements on attacking graph learning models, there are several limitations remain to be overcome:
\begin{itemize}
	\item \textbf{Unnoticeability}. Most works are unaware of preserving the adversarial attacks from noticeable impacts, they simply consider the lower attack budgets but far from enough instead.
	\item \textbf{Scalability}. Existing works are mainly focus on a relatively small-scale graph, however, million-scale or even larger graphs are commonly seen in real life, and efficient attacks on larger graph are leaving unexplored.\footnote{There has been some efforts on large-scale graph computation that would be useful for graph learning methods. \cite{zhu2016gemini,yang2019knightking}}
	\item \textbf{Knowledge}. It is common to assume that attackers have \emph{perfect knowledge} about the dataset, but it is unreasonable and impractical due to the limited access of attackers. Nevertheless, very few works conduct attacks with moderate or even minimal knowledge.
	\item \textbf{Physical Attack}. Most of the existing works conduct attacks on ideal datasets. However, in real world, attacks need to consider more factors. For instance, in a physical attack the adversarial examples as input will be distorted accidentally and it often fails to achieve the desired results. Unlike conducting attacks on ideal datasets, this brings more challenges to attackers.
\end{itemize}

\section{Defense}
\label{defense}
The proposed attack methods have made researchers realize the importance of the robustness of deep learning models. Relatively, some defense methods have also been proposed. In this section, we will give some general definitions of defense models against adversarial attack methods on graph data and its related concepts. In addition, this section systematically classifies existing defense methods and details some typical defense algorithms.

\subsection{Definition}
\label{defenseDefinition}
Simply put, the purpose of defense is to make the performance of the model still maintain a certain stability on the data that is maliciously disturbed. Although some defense models have been proposed, there is no clear and unified definition of the defense problem. To facilitate the discussion of the following, we propose a unified formulation of the defense problem.
\nosection{Defense on Graph Data} Most symbols are the same as mentioned in Section \ref{attack}, and we define $\Tilde{f}$ as a deep learning function with the loss function $\mathcal{\Tilde{L}}$ designed for defense, it receives a graph either perturbed or not. Then the defense problem can be defined as:

\begin{equation}
	\begin{split}
		& \underset{\hat{G} \in {\Psi(G)}\cup{\mathcal{G}}}{\text{minimize}} \sum_{t_i \in \mathcal{T}} \mathcal{\Tilde{L}}(\Tilde{f}_{\theta^{\ast}}(\hat{G}^{t_i}, X, t_i), y_i)  \\
		& s.t. \ \theta^{\ast} = \mathop{\arg \min}_{\theta} \sum_{v_j \in \mathcal{S}_L}\mathcal{\Tilde{L}}(\Tilde{f}_{\theta}(\tilde{G}^{v_j}, X, v_j), y_j)
	\end{split}
\end{equation}

where $\tilde{G} \in {\Psi(\mathcal{G})}\cup{\mathcal{G}}$ can be a well-designed graph $\hat{G}$ for the purpose of defense, or a clean graph $G$, which depends on whether the model have been attacked.

\subsection{Taxonomies of Defenses}
In this section, we divide the existing defense methods into several categories according to the used algorithms and describe them by examples. To the best of our knowledge, it's the first time for those defense methods to be clearly classified and the first time for those types to be clearly defined. All taxonomies are listed  below:

\nosection{\textbf{Preprocessing-based Defense}} As the most intuitive way, directly manipulating the raw data has a great impact on the model performance. Additionally, preprocessing raw data is independent to model structures and training methods, which gives considerable scalability and transferability. Some existing works\cite{wu2019adversarial} improve model robustness by conducting certain preprocessing steps before training, and we define this type of defense methods as \emph{Preprocessing-based Defense}. In addition, Wu et al. \cite{wu2019adversarial} try to drop edges that connect nodes with low similarity score, which could reduce the risk for those edges of being attack and nearly does no harm to the model performance.

\nosection{\textbf{Structure-based Defense}} In addition to raw data, model structure is also crucial to the model performance. Some existing works modify the model structure, such as GCN, to gain more robustness, and we define this type of defense methods as \emph{Structure-based Defense}. Instead of GCN's graph convolutional layers, Zhu et al. \cite{zhu2019robust} use Gaussian-based graph convolutional layers, which learns node embeddings from Gaussian distributions and assign attention weight according to their variances. Wang et al. \cite{wang2019adversarial} propose dual-stage aggregation as a Graph Neural Network (GNN) encoder layer to learn the information of neighborhoods, and adopt GAN \cite{goodfellow2014generative} to conduct contrastive learning. Tang et al. \cite{tang2019robust} initialize the model by meta-optimization and penalize the aggregation process of GNN. And Ioannidis et al. \cite{ioannidis2019edge} propose a novel GCN architecture, named Adaptive Graph Convolutional Network (AGCN), to conduct robust semi-supervised learning.

\nosection{\textbf{Adversarial-based Defense}} Adversarial training has been widely used in deep learning due to its excellent performance. Some researchers successfully adopt adversarial training from other fields into graph domain to improve model robustness, and we define the defense methods which use adversarial training as \emph{Adversarial-based Defense}. There are two types of adversarial training: i) \emph{Training with adversarial goals}. Some adversarial training methods gradually optimize the model in a continuous min-max way, under the guide of two opposite (minimize and maximize) objective functions \cite{xu2019topology, sun2019virtual, feng2019graph}; ii) \emph{Training with adversarial examples}. Other adversarial-based models are fed with adversarial samples during training, which helps the model learn to adjust to adversarial samples and thus reduces the negative impacts of those potential attack samples \cite{chen2019can, wang2019graphdefense, deng2019batch, he2018adversarial}.

\nosection{\textbf{Objective-based Defense}} As a simple and effective method, modifying objective function plays an important role in improving the model robustness. Many existing works attempt to train a robust model against adversarial attacks by optimizing the objective function, and we define this type of defense methods as \emph{Objective-based Defense}. However, there is a little intersection between the definition of \emph{Objective-based Defense} and \emph{Adversarial-based Defense} mentioned above, because the min-max adversarial training (the first type of \emph{Adversarial-based Defense}) is also in the range of objective optimization. Therefore, to accurately distinguish the definition boundary between \emph{Adversarial-based Defense} and \emph{Objective-based Defense}, we only consider those methods which are objective-based but not adversarial-based as the instances of \emph{Objective-based Defense}. Z\"ugner et al. \cite{zugner2019certifiable} and Bojchevski et al. \cite{bojchevski2019certifiable} combine hinge loss and cross entropy loss to perform a robust optimization. Jin et al. \cite{jin2019power} and Chen et al. \cite{chen2019can} regularize the training process by studying the characteristics of graph powering and smoothing the cross-entropy loss function, respectively.

\nosection{\textbf{Detection-based Defense}} Some existing works focus on the detection of adversarial attacks, or the certification of model/node robustness, which we define as \emph{Detection-based Defense}. Although those detection-based methods are unable to improve the model robustness directly, they can serve as supervisors who keep monitoring the model security and alarm for awareness when an attack is detected. Z\"ugner et al. \cite{zugner2019certifiable} and Bojchevski et al. \cite{bojchevski2019certifiable} propose novel methods to certificate whether a node is robust, and a robust node means it won't be affected even under the worse-case/strongest attack. Pezeshkpour et al. \cite{pezeshkpour2019investigating} study the impacts of adding/removing edges by performing adversarial modifications on the graph. Xu et al. \cite{xu2018characterizing} adopt link prediction and its variants to detect the potential risks, and for example, link with low score could be a maliciously added edge. Zhang et al. \cite{zhang2019comparing} utilize perturbations to explore the model structure and propose a method to detect adversarial attack. Hou et al. \cite{hou2019alphacyber} detect the target node by uncovering the poisoning nodes injected in the heterogeneous graph \cite{sun2011pathsim}. Ioannidis et al. \cite{ioannidis2019graphsac} effectively detect anomalous nodes in large-scale graphs by applying a graph-based random sampling and consensus strategies.

\nosection{\textbf{Hybrid Defense}} As the name suggested, we define \emph{Hybrid Defense} to denote the defense method which consists of two or more types of different defense algorithms mentioned above. Many researches flexibly combine several types of defense algorithms to achieve better performance, thus alleviate the limitations of only using a single method. As mentioned above, Z\"ugner et al. \cite{zugner2019certifiable} and Bojchevski et al. \cite{bojchevski2019certifiable} address the certification problem of node robustness and conduct robust training with objective-based optimization. Wang et al. \cite{wang2019adversarial} focus on improving the model structure and adversarially train the model by GAN. Chen et al. \cite{chen2019can} adopt adversarial training and other regularization mechanisms (e.g., gradient smoothing, loss smoothing). Xu et al. \cite{xu2018characterizing} propose a novel graph generation method together with other detection mechanisms (e.g., link prediction, subgraph sampling, outlier detect) as preprocessing to detect potential malicious edges. Miller et al. \cite{miller2019improving} consider a combination of the features come from graph structure and origin node attributes and use well-designed training data selection methods to do a classification task. These are instances of the hybrid defense.

\nosection{\textbf{Others}} Currently, the amount of researches for defense is far less than that of attack on the graph domain. To the best of our knowledge, most existing works for defense only focus on node classification tasks, and there are a lot of opportunities to study defense methods with different tasks on graph, which will enrich our defense types. For example, Pezeshkpour et al. \cite{pezeshkpour2019investigating} evaluate the robustness of several link prediction models and Zhou et al. \cite{zhou2019adversarial} adopt the heuristic approach to reduce the damage of attacks.

\begin{table*}[]

	\centering
	\caption{Related works on defense in details. The type is divided according to the algorithm.}
	\label{defense_table}
	\scalebox{0.8}{
		\begin{tabular}{c|cccccccccc}
			\hline
			\hline
			Reference                   &
			Model                       &
			Algorithm                   &
			Type                        &
			Target Task                 &
			Target Model                &
			Baseline                    &
			Metric                      &
			Dataset                       \\

			\hline
			\begin{tabular}[c]{@{}c@{}} \cite{zugner2019certifiable} \end{tabular} &
			\begin{tabular}[c]{@{}c@{}} GNN (trained with\\RH-U) \end{tabular} &
			\begin{tabular}[c]{@{}c@{}} Robustness certification,\\Objective based \end{tabular} &
			\begin{tabular}[c]{@{}c@{}} Hybrid \end{tabular} &
			\begin{tabular}[c]{@{}c@{}} Node\\classification \end{tabular} &
			\begin{tabular}[c]{@{}c@{}} GNN,\\GCN \end{tabular} &
			\begin{tabular}[c]{@{}c@{}} GNN\\(trained with\\CE, RCE, RH) \end{tabular} &
			\begin{tabular}[c]{@{}c@{}} Accuracy,\\Averaged\\Worst-case\\Margin \end{tabular} &
			\begin{tabular}[c]{@{}c@{}} Citeseer,\\Cora-ML,\\Pubmed \end{tabular}   \\
			\hline

			\begin{tabular}[c]{@{}c@{}} \cite{xu2019topology} \end{tabular} &
			\begin{tabular}[c]{@{}c@{}} - \end{tabular} &
			\begin{tabular}[c]{@{}c@{}} Adversarial training \end{tabular} &
			\begin{tabular}[c]{@{}c@{}} Adversarial\\training \end{tabular} &
			\begin{tabular}[c]{@{}c@{}} Node\\classification \end{tabular} &
			\begin{tabular}[c]{@{}c@{}} GCN \end{tabular} &
			\begin{tabular}[c]{@{}c@{}} GCN \end{tabular} &
			\begin{tabular}[c]{@{}c@{}} Accuracy, \\Misclassification rate \end{tabular} &
			\begin{tabular}[c]{@{}c@{}} Citeseer,\\Cora \end{tabular}   \\
			\hline

			\begin{tabular}[c]{@{}c@{}} \cite{jin2019power} \end{tabular} &
			\begin{tabular}[c]{@{}c@{}} r-GCN,\\VPN \end{tabular} &
			\begin{tabular}[c]{@{}c@{}} Graph\\powering \end{tabular} &
			\begin{tabular}[c]{@{}c@{}} Objective\\based \end{tabular} &
			\begin{tabular}[c]{@{}c@{}} Node\\classification \end{tabular} &
			\begin{tabular}[c]{@{}c@{}} GCN \end{tabular} &
			\begin{tabular}[c]{@{}c@{}} ManiReg,\\ICA,\\Vanilla GCN,\\... \end{tabular} &
			\begin{tabular}[c]{@{}c@{}} Accuracy, Robustness merit,\\Attack deterioration \end{tabular} &
			\begin{tabular}[c]{@{}c@{}} Citeseer,\\Cora,\\Pubmed \end{tabular}   \\
			\hline

			\begin{tabular}[c]{@{}c@{}} \cite{wu2019adversarial} \end{tabular} &
			\begin{tabular}[c]{@{}c@{}} - \end{tabular} &
			\begin{tabular}[c]{@{}c@{}} Drop edges \end{tabular} &
			\begin{tabular}[c]{@{}c@{}} Preprocessing \end{tabular} &
			\begin{tabular}[c]{@{}c@{}} Node\\classification \end{tabular} &
			\begin{tabular}[c]{@{}c@{}} GCN \end{tabular} &
			\begin{tabular}[c]{@{}c@{}} GCN \end{tabular} &
			\begin{tabular}[c]{@{}c@{}} Accuracy, \\Classification margin \end{tabular} &
			\begin{tabular}[c]{@{}c@{}} Citeseer,\\Cora-ML,\\Polblogs \end{tabular}   \\
			\hline
			\begin{tabular}[c]{@{}c@{}} \cite{wang2019adversarial} \end{tabular} &
			\begin{tabular}[c]{@{}c@{}} DefNet \end{tabular} &
			\begin{tabular}[c]{@{}c@{}} GAN,\\GER,\\ACL \end{tabular} &
			\begin{tabular}[c]{@{}c@{}} Hybrid \end{tabular} &
			\begin{tabular}[c]{@{}c@{}} Node\\classification \end{tabular} &
			\begin{tabular}[c]{@{}c@{}} GCN,\\GraphSAGE \end{tabular} &
			\begin{tabular}[c]{@{}c@{}} GCN,\\GraphSAGE \end{tabular} &
			\begin{tabular}[c]{@{}c@{}} Classification\\margin \end{tabular} &
			\begin{tabular}[c]{@{}c@{}} Citeseer,\\Cora,\\Polblogs \end{tabular}   \\
			\hline
			\begin{tabular}[c]{@{}c@{}} \cite{pezeshkpour2019investigating} \end{tabular} &
			\begin{tabular}[c]{@{}c@{}} CRIAGE \end{tabular} &
			\begin{tabular}[c]{@{}c@{}} Adversarial\\modification \end{tabular} &
			\begin{tabular}[c]{@{}c@{}} Robustness\\evaluation \end{tabular} &
			\begin{tabular}[c]{@{}c@{}} Link\\prediction \end{tabular} &
			\begin{tabular}[c]{@{}c@{}} Knowledge\\graph\\embeddings \end{tabular} &
			\begin{tabular}[c]{@{}c@{}} - \end{tabular} &
			\begin{tabular}[c]{@{}c@{}} Hits@K,\\MRR \end{tabular} &
			\begin{tabular}[c]{@{}c@{}} Nations,\\Kinship,\\WN18,\\YAGO3-10 \end{tabular}   \\
			\hline
			\begin{tabular}[c]{@{}c@{}} \cite{zhu2019robust} \end{tabular} &
			\begin{tabular}[c]{@{}c@{}} RGCN \end{tabular} &
			\begin{tabular}[c]{@{}c@{}} Gaussian-based\\GCN \end{tabular} &
			\begin{tabular}[c]{@{}c@{}} Structure\\based \end{tabular} &
			\begin{tabular}[c]{@{}c@{}} Node\\classification \end{tabular} &
			\begin{tabular}[c]{@{}c@{}} GCN \end{tabular} &
			\begin{tabular}[c]{@{}c@{}} GCN,\\GAT \end{tabular} &
			\begin{tabular}[c]{@{}c@{}} Accuracy \end{tabular} &
			\begin{tabular}[c]{@{}c@{}} Citeseer,\\Cora,\\Pubmed \end{tabular}   \\
			\hline
			\begin{tabular}[c]{@{}c@{}} \cite{chen2019can} \end{tabular} &
			\begin{tabular}[c]{@{}c@{}} Global-AT,\\Target-AT,\\SD, SCEL \end{tabular} &
			\begin{tabular}[c]{@{}c@{}} Adversarial\\training,\\Smooth defense \end{tabular} &
			\begin{tabular}[c]{@{}c@{}} Hybrid \end{tabular} &
			\begin{tabular}[c]{@{}c@{}} Node\\classification \end{tabular} &
			\begin{tabular}[c]{@{}c@{}} GNN \end{tabular} &
			\begin{tabular}[c]{@{}c@{}} AT \end{tabular} &
			\begin{tabular}[c]{@{}c@{}} ADR,\\ACD \end{tabular} &
			\begin{tabular}[c]{@{}c@{}} Citeseer,\\Cora,\\PolBlogs \end{tabular}   \\
			\hline
			\begin{tabular}[c]{@{}c@{}} \cite{sun2019virtual} \end{tabular} &
			\begin{tabular}[c]{@{}c@{}} SVAT,\\DVAT \end{tabular} &
			\begin{tabular}[c]{@{}c@{}} VAT \end{tabular} &
			\begin{tabular}[c]{@{}c@{}} Adversarial\\training \end{tabular} &
			\begin{tabular}[c]{@{}c@{}} Node\\classification \end{tabular} &
			\begin{tabular}[c]{@{}c@{}} GCN \end{tabular} &
			\begin{tabular}[c]{@{}c@{}} GCN \end{tabular} &
			\begin{tabular}[c]{@{}c@{}} Accuracy \end{tabular} &
			\begin{tabular}[c]{@{}c@{}} Citeseer,\\Cora,\\Pubmed \end{tabular}   \\
			\hline
			\begin{tabular}[c]{@{}c@{}} \cite{zhang2019comparing} \end{tabular} &
			\begin{tabular}[c]{@{}c@{}} - \end{tabular} &
			\begin{tabular}[c]{@{}c@{}} KL\\divergence \end{tabular} &
			\begin{tabular}[c]{@{}c@{}} Detection\\based \end{tabular} &
			\begin{tabular}[c]{@{}c@{}} Node\\classification \end{tabular} &
			\begin{tabular}[c]{@{}c@{}} GCN,\\GAT \end{tabular} &
			\begin{tabular}[c]{@{}c@{}} - \end{tabular} &
			\begin{tabular}[c]{@{}c@{}} Classification margin,\\Accuracy, ROC, AUC \end{tabular} &
			\begin{tabular}[c]{@{}c@{}} Citeseer,\\Cora,\\PolBlogs \end{tabular}   \\
			\hline
			\begin{tabular}[c]{@{}c@{}} \cite{feng2019graph} \end{tabular} &
			\begin{tabular}[c]{@{}c@{}} GCN-GATV \end{tabular} &
			\begin{tabular}[c]{@{}c@{}} GAT,\\VAT \end{tabular} &
			\begin{tabular}[c]{@{}c@{}} Adversarial\\training \end{tabular} &
			\begin{tabular}[c]{@{}c@{}} Node\\classification \end{tabular} &
			\begin{tabular}[c]{@{}c@{}} GCN \end{tabular} &
			\begin{tabular}[c]{@{}c@{}} DeepWalk,\\GCN,\\GraphSGAN,\\... \end{tabular} &
			\begin{tabular}[c]{@{}c@{}} Accuracy \end{tabular} &
			\begin{tabular}[c]{@{}c@{}} Citeseer,\\Cora,\\NELL \end{tabular}   \\

			\hline
			\begin{tabular}[c]{@{}c@{}} \cite{deng2019batch} \end{tabular} &
			\begin{tabular}[c]{@{}c@{}} S-BVAT,\\O-BVAT \end{tabular} &
			\begin{tabular}[c]{@{}c@{}} BVAT \end{tabular} &
			\begin{tabular}[c]{@{}c@{}} Adversarial\\training \end{tabular} &
			\begin{tabular}[c]{@{}c@{}} Node\\classification \end{tabular} &
			\begin{tabular}[c]{@{}c@{}} GCN \end{tabular} &
			\begin{tabular}[c]{@{}c@{}} ManiReg,\\GAT,\\GPNN,\\GCN,\\VAT,\\... \end{tabular} &
			\begin{tabular}[c]{@{}c@{}} Accuracy \end{tabular} &
			\begin{tabular}[c]{@{}c@{}} Citeseer,\\Cora,\\Pubmed,\\NELL \end{tabular}   \\
			\hline
			\begin{tabular}[c]{@{}c@{}} \cite{tang2019robust} \end{tabular} &
			\begin{tabular}[c]{@{}c@{}} PA-GNN \end{tabular} &
			\begin{tabular}[c]{@{}c@{}} Penalized aggregation,\\Meta learning \end{tabular} &
			\begin{tabular}[c]{@{}c@{}} Structure\\based \end{tabular} &
			\begin{tabular}[c]{@{}c@{}} Node\\classification \end{tabular} &
			\begin{tabular}[c]{@{}c@{}} GNN \end{tabular} &
			\begin{tabular}[c]{@{}c@{}} GCN,\\GAT,\\PreProcess,\\RGCN,\\VPN \end{tabular} &
			\begin{tabular}[c]{@{}c@{}} Accuracy \end{tabular} &
			\begin{tabular}[c]{@{}c@{}} Pubmed,\\Reddit,\\Yelp-Small,\\Yelp-Large \end{tabular}   \\
			\hline
			\begin{tabular}[c]{@{}c@{}} \cite{wang2019graphdefense} \end{tabular} &
			\begin{tabular}[c]{@{}c@{}} GraphDefense \end{tabular} &
			\begin{tabular}[c]{@{}c@{}} Adversarial\\training \end{tabular} &
			\begin{tabular}[c]{@{}c@{}} Adversarial\\training \end{tabular} &
			\begin{tabular}[c]{@{}c@{}} Node/Graph\\classification \end{tabular} &
			\begin{tabular}[c]{@{}c@{}} GCN \end{tabular} &
			\begin{tabular}[c]{@{}c@{}} Drop edges,\\Discrete AT \end{tabular} &
			\begin{tabular}[c]{@{}c@{}} Accuracy \end{tabular} &
			\begin{tabular}[c]{@{}c@{}} Cora,\\Citeseer,\\Reddit \end{tabular}   \\
			\hline
			\begin{tabular}[c]{@{}c@{}} \cite{hou2019alphacyber} \end{tabular} &
			\begin{tabular}[c]{@{}c@{}} HG-HGC \end{tabular} &
			\begin{tabular}[c]{@{}c@{}} HG-Defense \end{tabular} &
			\begin{tabular}[c]{@{}c@{}} Detection\\based \end{tabular} &
			\begin{tabular}[c]{@{}c@{}} Malware detection \end{tabular} &
			\begin{tabular}[c]{@{}c@{}} Malware\\detection\\system \end{tabular} &
			\begin{tabular}[c]{@{}c@{}} Other malware\\detection systems \end{tabular} &
			\begin{tabular}[c]{@{}c@{}} Detection rate \end{tabular} &
			\begin{tabular}[c]{@{}c@{}} Tencent Security \\Lab Dataset \end{tabular}   \\
			\hline
			\begin{tabular}[c]{@{}c@{}} \cite{ioannidis2019edge} \end{tabular} &
			\begin{tabular}[c]{@{}c@{}} AGCN \end{tabular} &
			\begin{tabular}[c]{@{}c@{}} Adaptive GCN with\\edge dithering\end{tabular} &
			\begin{tabular}[c]{@{}c@{}} Structure\\based \end{tabular} &
			\begin{tabular}[c]{@{}c@{}} Node\\classification \end{tabular} &
			\begin{tabular}[c]{@{}c@{}} GCN \end{tabular} &
			\begin{tabular}[c]{@{}c@{}} GCN \end{tabular} &
			\begin{tabular}[c]{@{}c@{}} Accuracy \end{tabular} &
			\begin{tabular}[c]{@{}c@{}} Citeseer, \\PolBlogs, \\Cora, Pubmed \end{tabular}   \\
			\hline
			\begin{tabular}[c]{@{}c@{}}	\cite{ioannidis2019graphsac} \end{tabular} &
			\begin{tabular}[c]{@{}c@{}} GraphSAC \end{tabular} &
			\begin{tabular}[c]{@{}c@{}} Random sampling, \\Consensus \end{tabular} &
			\begin{tabular}[c]{@{}c@{}} Detection\\based \end{tabular} &
			\begin{tabular}[c]{@{}c@{}} Anomaly \\detection \end{tabular} &
			\begin{tabular}[c]{@{}c@{}} Anomaly \\model\end{tabular} &
			\begin{tabular}[c]{@{}c@{}} GAE,\\Degree,\\Cut ratio,\\... \end{tabular} &
			\begin{tabular}[c]{@{}c@{}} AUC \end{tabular} &
			\begin{tabular}[c]{@{}c@{}} Citeseer,\\PolBlogs,\\Cora, Pubmed \end{tabular}   \\
			\hline
			\begin{tabular}[c]{@{}c@{}}	\cite{bojchevski2019certifiable} \end{tabular} &
			\begin{tabular}[c]{@{}c@{}} GNN (train with \\ $L_{RCE}$, $L_{CEM}$) \end{tabular} &
			\begin{tabular}[c]{@{}c@{}} Robustness certification,\\Objective based \end{tabular} &
			\begin{tabular}[c]{@{}c@{}} Hybrid \end{tabular} &
			\begin{tabular}[c]{@{}c@{}} Node\\classification \end{tabular} &
			\begin{tabular}[c]{@{}c@{}} GNN \end{tabular} &
			\begin{tabular}[c]{@{}c@{}} GNN \end{tabular} &
			\begin{tabular}[c]{@{}c@{}} Accuracy,\\Worst-case\\Margin \end{tabular} &
			\begin{tabular}[c]{@{}c@{}} Cora-ML,\\Citeseer,\\Pubmed \end{tabular}   \\
			\hline
			\begin{tabular}[c]{@{}c@{}} \cite{zhou2019adversarial} \end{tabular} &
			\begin{tabular}[c]{@{}c@{}} IDOpt,\\IDRank \end{tabular} &
			\begin{tabular}[c]{@{}c@{}} Integer Program,\\Edge Ranking \end{tabular} &
			\begin{tabular}[c]{@{}c@{}} Heuristic\\algorithm \end{tabular} &
			\begin{tabular}[c]{@{}c@{}} Link\\prediction \end{tabular} &
			\begin{tabular}[c]{@{}c@{}} Similarity-based \\link prediction\\ models \end{tabular} &
			\begin{tabular}[c]{@{}c@{}} PPN \end{tabular} &
			\begin{tabular}[c]{@{}c@{}} DPR \end{tabular} &
			\begin{tabular}[c]{@{}c@{}} PA,\\PLD,\\TVShow,\\Gov \end{tabular}   \\
			\hline
			\begin{tabular}[c]{@{}c@{}} \cite{miller2019improving} \end{tabular} &
			\begin{tabular}[c]{@{}c@{}} SVM with\\a radial basis\\function kernel \end{tabular} &
			\begin{tabular}[c]{@{}c@{}} Augmented feature,\\Edge selecting \end{tabular} &
			\begin{tabular}[c]{@{}c@{}} Hybrid \end{tabular} &
			\begin{tabular}[c]{@{}c@{}} Node\\classification \end{tabular} &
			\begin{tabular}[c]{@{}c@{}} SVM \end{tabular} &
			\begin{tabular}[c]{@{}c@{}} GCN \end{tabular} &
			\begin{tabular}[c]{@{}c@{}} Classification\\marigin \end{tabular} &
			\begin{tabular}[c]{@{}c@{}} Cora,\\Citeseer \end{tabular}   \\
			\hline
			\begin{tabular}[c]{@{}c@{}} \cite{he2018adversarial} \end{tabular} &
			\begin{tabular}[c]{@{}c@{}} APR,\\AMF \end{tabular} &
			\begin{tabular}[c]{@{}c@{}} MF-BPR\\based AT \end{tabular} &
			\begin{tabular}[c]{@{}c@{}} Adversarial\\training \end{tabular} &
			\begin{tabular}[c]{@{}c@{}} Recommendation \end{tabular} &
			\begin{tabular}[c]{@{}c@{}} MF-BPR \end{tabular} &
			\begin{tabular}[c]{@{}c@{}} ItemPop,\\MF-BPR,\\CDAE,\\NeuMF,\\IRGAN \end{tabular} &
			\begin{tabular}[c]{@{}c@{}} HR,\\NDCG \end{tabular} &
			\begin{tabular}[c]{@{}c@{}} Yelp,\\Pinterest,\\Gowalla \end{tabular}   \\
			\hline
			\begin{tabular}[c]{@{}c@{}} \cite{xu2018characterizing} \end{tabular} &
			\begin{tabular}[c]{@{}c@{}} SL, OD,\\P+GGD,\\ENS,GGD \end{tabular} &
			\begin{tabular}[c]{@{}c@{}} Link prediction,\\Subsampling,\\Neighbour analysis \end{tabular} &
			\begin{tabular}[c]{@{}c@{}} Hybrid \end{tabular} &
			\begin{tabular}[c]{@{}c@{}} Node\\classification \end{tabular} &
			\begin{tabular}[c]{@{}c@{}} GNN, GCN \end{tabular} &
			\begin{tabular}[c]{@{}c@{}} LP \end{tabular} &
			\begin{tabular}[c]{@{}c@{}} AUC \end{tabular} &
			\begin{tabular}[c]{@{}c@{}} Citeseer,\\Cora \end{tabular}   \\
			\hline
			\hline
		\end{tabular}
	}
\end{table*}

\subsection{Summary: Defense on Graph}
In this section, we will introduce the contributions and limitations of current works about defense. Then, we will discuss the potential research opportunities in this area. The details and comparisons of various methods are placed in Table \ref{defense_table}.

\subsubsection{Major Contributions}
Currently, most of the methods to improve the robustness of GNNs start from two aspects: a robust training method or a robust model structure. Among them, the training methods are mostly adversarial training, and many of the model structure improvements are made using the attention mechanism. In addition, there are some studies that do not directly improve the robustness of GNNs but try to verify the robustness or try to detect the data that is disturbed. In this part, considering the different ways in which existing methods contribute to the adversarial defense of graphs, we summarize the contributions of graph defense from the following three perspectives: adversarial learning, model improvement and others.
\nosection{\textbf{Adversarial Learning}}
As a successful method that has shown to be effective in defending the adversarial attacks of image data and test data, adversarial learning \cite{goodfellow2014explaining} augments the training data with adversarial examples during the training stage. Some researchers have also tried to apply this kind of idea to graph defense methods. Wang et al.
\cite{wang2019adversarial} think the vulnerabilities of graph neural networks are related to the aggregation layer and the perceptron layer. To address these two disadvantages, they propose an adversarial training framework with a modified GNN model to improve the robustness of GNNs.
Chen et al. \cite{chen2019can} propose different defense strategies based on adversarial training for target and global adversarial attack with smoothing distillation and smoothing cross-entropy loss function.
Feng et al. \cite{feng2019graph} propose a method of adversarial training for the attack on node features with a graph adversarial regularizer which encourages the model to generate similar predictions on the perturbed target node and its connected nodes.
Sun et al. \cite{sun2019virtual} successfully transfer the efficiency of Virtual Adversarial Training (VAT) to the semi-supervised node classification task on graphs and applies some regularization mechanisms on original GCN to refine its generalization performance.
Wang et al. \cite{wang2019graphdefense} point out that the values of perturbation in adversarial training could be continuous or even negative. And they propose an adversarial training method to improve the robustness of GCN.
He et al. \cite{he2018adversarial} adopt adversarial training to Bayesian Personalized Ranking (BPR) \cite{rendle2009bpr} on recommendation by adding adversarial perturbations on embedding vectors of the user and item.
\nosection{\textbf{Model Improvement}}
In addition to improvements in training methods, many studies have focused on improving the model itself. In recent years, the attention mechanism \cite{vaswani2017attention} has shown extraordinary performance in the field of natural language processing. Some studies of graph defense have borrowed the way that they can automatically give weight to different features to reduce the influence of perturbed edges on features.
Zhu et al. \cite{zhu2019robust} use Gaussian distributions to absorb the effects of adversarial attacks and introduce a variance-based attention mechanism to prevent the propagation of adversarial attacks.
Tang et al. \cite{tang2019robust} propose a novel framework base on a penalized aggregation mechanism which restrict the negative impact of adversarial edges.
Hou et al. \cite{hou2019alphacyber} enhance the robustness of malware detection system in Android by their well-designed defense mechanism to uncover the possible injected poisoning nodes.
Jin et al. \cite{jin2019power} point out the basic flaws of the Laplacian operator in origin GCN and propose a variable power operator to alleviate the issues.
Miller et al. \cite{miller2019improving} propose to use unsupervised methods to extract the features of the graph structure and use support vector machines to complete the task of node classification. In addition, they also propose two new training data partitioning strategies.
\nosection{\textbf{Others}}
In addition to improving the robustness of the model, there are also some studies that have made other contributions around robustness, such as detecting disturbed edges, certificating node robustness, analyses of current attack methods and so on.
Z\"ugner et al. \cite{zugner2019certifiable} propose the first work on certifying the robustness of GNNs. The method can give robustness certification which states whether a node is robust under a certain space of perturbations.
Zhang et al. \cite{zhang2019comparing} study the defense methods' performance under random attacks and Nettack concluded that graph defense models which use structure exploration are more robust in general. Otherwise, this paper proposes a method to detect the perturbed edges by calculating the mean of the KL divergences \cite{kullback1951information} between the softmax probabilities of the node and it's neighbors.
Pezeshkpour et al. \cite{pezeshkpour2019investigating} introduce a novel method to automatically detect the error for knowledge graphs by conducting adversarial modifications on knowledge graphs. In addition, the method can study the interpretability of knowledge graph representations by summarizing the most influential facts for each relation.
Wu et al. \cite{wu2019adversarial} argue that the robustness issue is rooted in the local aggregation in GCN by analyzing attacks on GCN and propose an effective defense method based on preprocessing.
Bojchevski et al. \cite{bojchevski2019certifiable} propose a robustness certificate method that can certificate the robustness of GNNs regarding perturbations of the graph structure and the label propagation. In addition, they also propose a new robust training method guided by their own certificate  method.
Zhou et al. \cite{zhou2019adversarial} model the problem of robust link prediction as a Bayesian Stackelberg game \cite{von2010market} and propose two defense strategies to choose which link should be protected.
Ioannidis et al. \cite{ioannidis2019edge} propose a novel edge-dithering (ED) approach reconstructs the original neighborhood structure with high probability as the number of sampled graphs increases.
Ioannidis et al. \cite{ioannidis2019graphsac} introduce a graph-based random sampling and consensus approach to effectively detect anomalous nodes in large-scale graphs.

\subsubsection{Current Limitations}
As a new-rising branch that has not been sufficiently studied, current defense methods have several major limitations as follows:
\begin{itemize}
	\item \textbf{Diversity}. At present, most works of defense mainly focus on node classification tasks only. From the perspective of defender, they need to improve their model robustness on different tasks.
	\item \textbf{Scalability}. Whether a defense model can be widely used in practice largely depends on the cost of model training. Most existing works lack the consideration of training costs.
	\item \textbf{Theoretical Proof}. Most of current methods only illustrate their effectiveness by showing experimental results and textual descriptions. It will be great if a new robust method's effectiveness can be proved theoretically.
\end{itemize}

\section{Metrics}
\label{metrics}
In this section, we first introduce the metrics that are common in graph analysis tasks, which are used in attack and defense scenarios as well. Next, we introduce some new metrics proposed in attack and defense works from three perspectives: effectiveness, efficiency, and imperceptibility.
\subsection{Common Metrics}
\begin{itemize}
	\item \textbf{FNR} and \textbf{FPR}. In the classification or clustering tasks, here is a category of metrics based on False Positive (\textit{FP}), False Negative (\textit{FN}), True Positive (\textit{TP}) and True Negative (\textit{TN}). In attack and defense scenarios, commonly used to quantify are False Negative Rate (\textit{FNR}) and False Positive Rate (\textit{FPR}). Specifically, \textit{FNR} is equal to \textit{FN} over (\textit{TP} + \textit{FN}), and \textit{FPR} is equal to \textit{FP} over (\textit{FP} + \textit{TN}). For all negative samples, the former describes the proportion of false positive samples detected, while the latter describes the proportion of false negative samples detected. In other words, for negative samples, the former is the error rate, and the latter is the miss rate.
	\item \textbf{Accuracy (Acc).} The Accuracy is one of the most commonly used evaluation metrics, which measures the quality of results based on the percentage of correct predictions over total instances.
	\item \textbf{F1-score.} The F1 score \cite{bojchevski2019adversarial} can be regarded as a harmonic average of model Precision and Recall score \cite{powers2011evaluation}, with a maximum value of 1 and a minimum value of 0.
	\item \textbf{Area Under Curve (AUC).} The AUC is the area under the Receiver Operating Characteristic (ROC) curve \cite{zhang2019comparing}. Different models corresponding to different ROC curves, and it is difficult to compare which one is better if there is a crossover between the curves, thus comparison based on the AUC score is more reasonable. In general, AUC indicates the probability that the predicted positive examples rank in front of the negative examples.

	\item \textbf{Average Precision (AP).} As the area under the Precision-Recall curve, AP is described as one of the most robust metrics \cite{boyd2013area}. Precision-Recall curve depicts the relationships between Precision and Recall. A good model should improve the Recall while preserving the Precision a relatively high score. In contrast, weaker models may lose more Precision in order to improve Recall. Comparing with the Precision-Recall curve, AP can show the performance of the model  more intuitively.

	\item \textbf{Mean Reciprocal Rank (MRR).} The MRR is a commonly used metric to measure a rank model. For a target query, if the first correct item is ranked $n_{th}$, then the MRR score is $1 / n$, and once there is no match, the score is 0. The MRR of the model is the sum of the scores of all queries.

	\item \textbf{Hits@K.} By calculating the rank (e.g., MRR) of all the ground-truth triples, Hits@K is the proportion of correct entities ranked in top K.

	\item \textbf{Modularity.} The Modularity is an important measure based on the assortative mixing \cite{newman2003mixing},  which usually used to assess the quality of different division for a particular network, especially the community structure is unknown \cite{newman2004finding}.

	\item \textbf{Normalized Mutual Information (NMI).} The NMI is another commonly used evaluation to measure the quality of clustering results, so as to analyze the network community structure \cite{danon2005comparing}. NMI further indicates the similarity between two partitions with mutual information and information entropy, while a larger value means a higher similarity.

	\item \textbf{Adjusted Rand Index (ARI).} The ARI is a measure of the similarity between two data clusterings: one given by the clustering process and the other defined by external criteria \cite{santos2009use}. Such a correction establishes a baseline by using the expected similarity of all pair-wise comparisons between clusterings specified by a random model. ARI measures the relation between pairs of dataset elements without labels' information, which can cooperate with conventional performance measures to detect classification algorithm. In addition, ARI can also use labels information for feature selection \cite{santos2009use}.

\end{itemize}

\subsection{Specific Metrics for Attack and Defense}
In order to measure the performance, a number of specific metrics for attack and defense appear in literature. Next, we will organize these metrics from three perspectives: effectiveness, efficiency, and imperceptibility.

\subsubsection{Effectiveness-relevant Metrics}
Both attack and defense require metrics to measure the performance of the target model before and after the attack. Therefore, we have summarized some metrics for measuring effectiveness and list them below:

\begin{itemize}
	\item \textbf{Attack Success Rate (ASR)}. ASR is the ratio of targets which will be successfully attacked within a given fixed budget \cite{chen2018link}. Correspondingly, we can conclude the formulation of ASR:
	      \begin{equation}
		      \text{ASR}=\frac{\text {Number  of  successful  attacks }}{\text {Number  of  attacks }}
	      \end{equation}

	\item \textbf{Classification Margin (CM)}. CM is only designed for the classification task. Under this scenario, attackers aim to perturbe a graph that misclassifies the target node and has maximal ``distance'' (in terms of log-probabilities/logits) to the correct class \cite{zugner2018adversarial}. Based on this, CM is well formulated as:
	      \begin{equation}
		      \text{CM}=\max_{y'_t \neq y_t} Z_{t,y'_t}-Z_{t,y_t}
	      \end{equation}
	      where $Z_t$ is the output of the target model with respect to node $t$, and $y_t \in \mathcal{C}$ is the \emph{correct} class label of $v_t$ while $y'_t$ is the \emph{wrong} one.

	\item \textbf{Averaged Worst-case Margin (AWM)}. Worst-case Margin (WM) is the minimum of the Classification Margin (CM) \cite{zugner2019certifiable}, and the average of WM is dynamically calculated over a mini-batch of nodes during training. The $B_s$ in the following equation denotes batch size.
	      \begin{equation}
		      \text{AWM} = \frac{1}{B_s} \sum^{i=B_s}_{i=1} \text{CM}_{\text{worst}_i}
	      \end{equation}

	\item \textbf{Robustness Merit (RM)}. It evaluates the robustness of the model by calculating the difference between the post-attack accuracy and pre-attack accuracy on GNNs \cite{jin2019power}. And $\mathcal{V^\text{attacked}}$ denotes the set of nodes theoretically affected by the attack.
	      \begin{equation}
		      \text{RM}=\text{Acc}_{V^{\text{attacked}}}^{\text{post-attacked}}-\text{Acc}^{\text{pre-attacked}}_{V^{\text{attacked}}}
	      \end{equation}

	\item \textbf{Attack Deterioration (AD)}. It evaluates the attack effect of the model prediction. Note that any added/dropped edges can only affect the nodes within the spatial scope in the origin network \cite{jin2019power}, due to the spatial scope limitaions of the GNNs.
	      \begin{equation}
		      \text{AD}=1-\frac{\text{Acc}^{\text{post-attacked}}_{\mathcal{V}^{\text{attacked}}}}{\text{Acc}^{\text{pre-attacked}}_{\mathcal{V}^{\text{attacked}}}}
	      \end{equation}

	\item \textbf{Average Defense Rate (ADR)}. ADR is the ratio of the difference between the ASR of attack on GNNs with and without defense, versus the ASR of attack on GNNs without defense \cite{chen2019can}. The higher ADR the better defense performance.
	      \begin{equation}
		      \text{ADR}=\frac{\text{ASR}^{\text{with-defense}}_{\mathcal{V}^{\text{attacked}}}}{\text{ASR}^{\text{without-defense}}_{\mathcal{V}^{\text{attacked}}}}-1
	      \end{equation}

	\item \textbf{Average Confidence Different (ACD)}. ACD is the average confidence different of nodes in $N_s$ before and after attack \cite{chen2019can}, where $N_s$ is the set of nodes which are classified correctly before attack in test set. Note that Confidence Different (CD) is equivalent to Classification Margin (CM).
	      \begin{equation}
		      \text{ACD}=\frac{1}{N_s}\sum_{v_{i}\in N_{s}} \text{CD}_{i}(\hat{A}_{v_{i}})-\text{CD}_{i}(A)
	      \end{equation}

	\item \textbf{Damage Prevention Ratio (DPR)}. Damage prevention is defined to measure the amount of damage that can be prevented by defense \cite{zhou2019adversarial}. Let $L_0$ be the defender's loss without attack, $L_A$ be the defender's loss under a certain attack strategy, $A$ and $L_D$ be the defender's loss with a certain defense strategy $D$. A better defense strategy leads to a larger DPR.
	      \begin{equation}
		      \text{DPR}_A^D=\frac{L_A-L_D}{L_A-L_0}
	      \end{equation}
\end{itemize}

\subsubsection{Efficiency-relevant Metrics}
Here we introduce some efficiency metrics which measure the cost of the attack and defense. For example, the metric of quantifying how much perturbations are required for the same effect.

\begin{itemize}
	\item \textbf{Average Modified Links (AML)}. AML is designed for the topology attack, which indicates the average perturbation size leading to a successful attack \cite{chen2018link}. Assume that attackers have limited budgets to attack a target model, the modified links (added or removed) are accumulated until attackers achieve their goal or run out of the budgets. Based on this, we can conclude the formulation of AML:
	      \begin{equation}
		      \text{AML}=\frac{\text {Number  of  modified  links }}{\text {Number  of  attacks }}
	      \end{equation}

\end{itemize}

\subsubsection{Imperceptibility-relevant Metrics}
Several metrics are proposed to measure the scale of the manipulations caused by attack and defense, which are summarized as follows:
\begin{itemize}
	\item \textbf{Similarity Score}. Generally speaking, Similarity Score is a measure to infer the likelihoods of the existence of links, which usually applied in the link-level task \cite{lu2011link,zhou2019attacking}. Specifically, suppose we want to figure out whether a particular link $(u,v)$ exists in a network, Similarity Score can be used to quantify the topology properties of node $u$ and $v$ (e.g., common neighbors, degrees), and a higher Similarity Score indicates a greater likelihood of connection between this pair. Usually, Similarity Score could be measured by the cosine similarity matrix \cite{sun2018data}.

	\item \textbf{Test Statistic $\Lambda$.} $\Lambda$ takes advantage of the distribution of node degrees to measure the similarity between graphs. Specifically,  Z\"ugner et al. \cite{zugner2018adversarial} state that multiset of node degrees $\mathit{d}_{G}$ in the graph $G$ follow the power-law like distribution, and they provide a method to approximate the main parameter $\alpha_{G}$ of the distribution. Through $\alpha_{G}$ and $\mathit{d}_{G}$, we can calculate $G$'s log-likelihood score $l(\mathit{d}_{G}, \alpha_{G})$. For the pre-attack graph $G_{pr}$ and post-attack graph $G_{po}$, we get ($\mathit{d}_{G_{pr}},\alpha_{G_{pr}}$) and ($\mathit{d}_{G_{po}}, \alpha_{G_{po}}$), respectively. Similarly, we can then define $\mathit{d}_{G_{q}} = \mathit{d}_{G_{pr}}  \cap \  \mathit{d}_{G_{po}}$ and estimate $\alpha_{G_{q}}$. The final test statistic of graphs can be formulated as:
	      \begin{equation}
		      \begin{aligned}
			      \Lambda(G_{pr}, G_{po}) & = 2 \cdot (-l(\mathit{d}_{G_{q}}, \alpha_{G_{q}}) \\
			                              & + l(\mathit{d}_{G_{po}}, \alpha_{G_{po}})
			      + l(\mathit{d}_{G_{pr}}, \alpha_{G_{pr}}))
		      \end{aligned}
	      \end{equation}
	      Finally, when the statistic $\Lambda$ satisfies a specified constraint, the model considers the perturbations are unnoticeable.

	\item \textbf{Attack Budget $\Delta$}. To ensure unnoticeable perturbations in the attack, previous work \cite{zugner2018adversarial} measures the perturbations in terms of the budget $\Delta$ and the test statistics $\Lambda$ w.r.t. log-likelihood score. More precisely, they accumulate the changes in node feature and the adjacency matrix, and limit it to a budge $\Delta$ to constrain perturbations. However, it is not suitable to deal with complicated situations \cite{zugner2018adversarial}.

	\item \textbf{Attack Effect.} Generally, the attack effect evaluates the impacts in the community detection task.  Given a confusion matrix $J$, where each element $J_{i,j}$ represents the number of shared nodes between original community and a perturbed one, the attack effect is simply defined as the accumulation of normalized entropy \cite{chen2019multiscale}. Suppose all the communities keep exactly the same after the attack, the attack effect will be equal to $0$.

\end{itemize}

\section{Open Problems}
\label{openProblems}
Graph adversarial learning has many problems worth to be studied by observing existing researches. In this section, we try to introduce several major research issues and discuss the potential solutions. Generally, we will discuss the open problems in three parts: attack, defense and evaluation metric.

\subsection{Attack}
We mainly approach the problems from three perspectives, attacks' side effects, performance, and prerequisites. First, we hope the inevitable disturbances in the attack can be stealthy and unnoticeable. Second, now that data is huge, attacks need to be efficient and scalable. Finally, the premise of the attack should not be too ideal, that is, it should be practical and feasible. Based on the limitations of previous works detailed in Section \ref{attack_limitation} and the discussion above, we rise several open problems in the following:

\begin{itemize}
	\item \textbf{Unnoticeable Attack}.  Adversaries want to keep the attack unnoticeable to avoid censorship. To this end, Z\"ugner et al. \cite{zugner2019adversarial} argue that the main property of the graph should be marginally changed after attack, e.g., attackers are restricted to maintain the degree distribution while attacking a specified graph, which is evaluated via a test static. Nevertheless, most of existing works ignore such a constraint despite achieving outstanding performance. To be more concealed, we believe that attackers should focus more on their perturbation impacts on a target graph, with more constraints placed on the attacks.
	\item \textbf{Efficient and Effective Algorithms}. Seeing from Table \ref{attack_table}, the most frequently used benchmark datasets are rather small-scale graphs, e.g., Cora \cite{mccallum2000automating} and Citeseer \cite{sen2008collective}. Currently, the majority of proposed methods are failed to attack a large-scale graph, for the reason that they need to store the entire information of the graph to compute the surrogate gradients or loss even with small changes, leading to the expensive computation and memory consumption. To address this problem, Z\"ugner et al. \cite{zugner2019adversarial} make an effort that derives an incremental update for all candidate set (potential perturbation edges) with a linear variant of GCN, which avoid the redundant computation while remain efficient attack performance. Unfortunately, the proposed method isn't suitable for a larger-scale graph yet. With the shortcoming that unable to conduct attacks on a larger-scale graph, a more efficient and effective attack method should be studied to address this practical problem. For instance, given a target node, the message propagation are only exits within its $k$-hops neighborhood.\footnote{Here $k$ depends on the specified method of message propagation and is usually of a smaller value, e.g., $k=2$ with a two-layer GCN.} In other words, we believe that adversaries can explore a heuristic method to attack an entire graph (large) via its represented subgraph (small), and naturally the complexity of time and space  are reduced.
	\item \textbf{Constraints on Attackers' Knowledge}. To conduct a real and practical black-box attack, attackers should not only be blind to the model knowledge, but also be strict to minimal data knowledge. This is a challenging setting from the perspective of attackers. Chen et al. \cite{chen2017practical} explore this setting by randomly sampling different size of subgraphs, showing that the attacks are failed with a small network size, and the performance increases as it gets larger. However, few works are aware of the constraints on attackers' knowledge, instead, they assume that perfect knowledge of the input data are available, and naturally perform well in attacking a fully ``exposed'' graph. Therefore, several works could be explored with such a strict constraint, e.g., attackers can train a substitute model on a certain part of the graph, learn the general patterns of conducting attacks on graph data, thus transfer to other models and entire graph with less prior knowledge.
	\item \textbf{Real-world Attack}. Attacks can't always be on an ideal assumption, i.e., studying attacks on a simple and static graphs. Such ideal, undistorted input is unrealistic in real cases. In other words, how can we improve existing models so that they can work in complex real-world environments?
\end{itemize}

\subsection{Defense}
Besides the open problems of attack mentioned above, there are still many interesting problems in defense on the graph, which deserve more attentions. Here we list some open problems that worth discussing:
\begin{itemize}

	\item \textbf{Defense on Various Tasks}. From table \ref{defense_table} we can see that, most existing works of defense \cite{zugner2019certifiable,xu2019topology,jin2019power,wu2019adversarial,wang2019adversarial} are focusing on the node classification task on graph, achieving promising performance. Except for node classification, there are various tasks on graph domain are important and should be pay more attention to. However, only a few works \cite{pezeshkpour2019investigating,zhou2019adversarial} try to investigate and improve model robustness on link prediction task on graph while a few works \cite{hou2019alphacyber,ioannidis2019graphsac,he2018adversarial} make an effort to defense on some tasks (e.g., malware detection, anomaly detection, recommendation) on other graph domains. Therefore, it is valuable to think about how to improve model robustness on various tasks or transfer the current defense methods to other tasks.

	\item \textbf{Efficient Defense.} Training cost is critical important factor to be taken into consideration under the industrial scenes. Therefore, it's worth to be studied how to guarantee acceptable cost while improving robustness. However, the currently proposed defense methods have rarely discussed the space-time efficiency of their algorithms. Wang et al. \cite{wang2019adversarial} design a bottleneck mapping to reduce the dimension of input for a better efficiency, but their experiments absence the consideration of large-scale graph. Z\"ugner et al. \cite{zugner2019certifiable} test the number of coverage iterations of their certification methods under the different number of nodes, but the datasets they used are still small. Wang et al. \cite{wang2019graphdefense} do not restrict the regularized adjacency matrix to be discrete during the training process, which improves the training efficiency.

	\item \textbf{Certification on Robustness}. Robustness of the graph model is always the main issue among all the existing works including attack and defense mentioned above. However, most researchers only focus on improving model (e.g., GCN) robustness, and try to prove the model robustness via the performance results of their model. The experimental results can indeed prove the robustness of the model to a certain extent. However, considering that the model performance is easily influenced by the hyperparameters, implementation method, random initialization, and even some hardware constraints (e.g., cpu, gpu, memory), it is difficult to guarantee that the promising and stable performance can be reobtained on different scenarios (e.g., datasets, tasks, parameter settings). There is a novel and cheaper way to prove robustness via certification that should be taken more seriously into consideration, which certificate the node's absolute robustness under arbitrary perturbations, but currently only a few works have paid attention to the certification of GNNs' robustness \cite{zugner2019certifiable,bojchevski2019certifiable} which are also a valuable research direction.

\end{itemize}

\subsection{Evaluation Metric}
As seen in Section \ref{metrics}, we got so many evaluation metrics in graph adversarial learning field, however, the number of effectiveness-relevant metrics are far more than the other two, which reflects the research emphasis on model performance is unbalanced. Therefore, we propose several potential works can be further studied:

\begin{itemize}
	\item \textbf{Measurement of Cost.} There are not many metrics to explore the model's efficiency which reflects the lack of research attention on it. The known methods roughly measure the cost via the number of modified edges which begs the question that is there another perspective to quantify the cost of attack and defense more precisely? In real world, the cost of adding an edge is rather different from removing one,\footnote{Usually, removing an edge is more expensive than adding one.} thus the cost between modified edges are unbalanced, which needs more reasonable evaluation metrics.

	\item \textbf{Measurement of Imperceptibility.} Different from the image data, where the perturbations are bounded in $\ell_p$-norm \cite{sun2018adversarial,goodfellow2014explaining,madry2017towards}, and could be used as an evaluation metric on attack impacts, but it is ill-defined on graph data. Therefore, how to better measure the effect of perturbations and make the attack more unnoticeable?
	\item \textbf{Appropriate Metric Selection.} With so many metrics proposed to evaluate the performance of the attack and defense algorithms, here comes a problem that how to determine the evaluation metrics in different scenarios?
\end{itemize}

\section{Conclusion}
\label{conclusion}
In this survey, we conduct a comprehensive review on graph adversarial learning, including attacks, defenses and corresponding evaluation metrics. To the best of our knowledge, this is the first work systemically summarize the extensive works in this field. Specifically, we present the recent developments of this area and introduce the arms race between attackers and defenders. Besides, we provide a reasonable taxonomy for both of them, and further give a unified problem formulation which makes it clear and understandable. Moreover, under the scenario of graph adversarial learning, we summarize and discuss the major contributions and limitations of current attack and defense methods respectively, along with the open problems of this area that worth exploring. Our works cover most of the relevant evaluation metrics in the graph adversarial learning field as well, aiming to provide a better understanding on these methods. In the future, we will measure the performance of proposed models with relevant metrics based on extensive empirical studies.

Hopefully, our works will serve as a reference and give researchers a comprehensive and systematical understanding of the fundamental issues, thus become a well starting point to study in this field.

\bibliographystyle{IEEEtran}
\bibliography{tpami22}

\begin{thebibliography}{10}
\providecommand{\url}[1]{#1}
\csname url@samestyle\endcsname
\providecommand{\newblock}{\relax}
\providecommand{\bibinfo}[2]{#2}
\providecommand{\BIBentrySTDinterwordspacing}{\spaceskip=0pt\relax}
\providecommand{\BIBentryALTinterwordstretchfactor}{4}
\providecommand{\BIBentryALTinterwordspacing}{\spaceskip=\fontdimen2\font plus
\BIBentryALTinterwordstretchfactor\fontdimen3\font minus
  \fontdimen4\font\relax}
\providecommand{\BIBforeignlanguage}[2]{{%
\expandafter\ifx\csname l@#1\endcsname\relax
\typeout{** WARNING: IEEEtran.bst: No hyphenation pattern has been}%
\typeout{** loaded for the language `#1'. Using the pattern for}%
\typeout{** the default language instead.}%
\else
\language=\csname l@#1\endcsname
\fi
#2}}
\providecommand{\BIBdecl}{\relax}
\BIBdecl

\bibitem{xiong2016achieving}
W.~Xiong, J.~Droppo, X.~Huang, F.~Seide, M.~Seltzer, A.~Stolcke, D.~Yu, and
  G.~Zweig, ``Achieving human parity in conversational speech recognition,''
  \emph{arXiv preprint arXiv:1610.05256}, 2016.

\bibitem{collobert2011natural}
R.~Collobert, J.~Weston, L.~Bottou, M.~Karlen, K.~Kavukcuoglu, and P.~Kuksa,
  ``Natural language processing (almost) from scratch,'' \emph{Journal of
  machine learning research}, vol.~12, no. Aug, pp. 2493--2537, 2011.

\bibitem{parkhi2015deep}
O.~M. Parkhi, A.~Vedaldi, A.~Zisserman \emph{et~al.}, ``Deep face
  recognition.'' in \emph{bmvc}, vol.~1, no.~3, 2015, p.~6.

\bibitem{krishnamurthy2018deep}
B.~Krishnamurthy and M.~Sarkar, ``Deep-learning network architecture for object
  detection,'' Dec.~11 2018, uS Patent 10,152,655.

\bibitem{papernot2017practical}
N.~Papernot, P.~McDaniel, I.~Goodfellow, S.~Jha, Z.~B. Celik, and A.~Swami,
  ``Practical black-box attacks against machine learning,'' in
  \emph{Proceedings of the 2017 ACM on Asia conference on computer and
  communications security}.\hskip 1em plus 0.5em minus 0.4em\relax ACM, 2017,
  pp. 506--519.

\bibitem{szegedy2013intriguing}
C.~Szegedy, W.~Zaremba, I.~Sutskever, J.~Bruna, D.~Erhan, I.~Goodfellow, and
  R.~Fergus, ``Intriguing properties of neural networks,'' \emph{arXiv preprint
  arXiv:1312.6199}, 2013.

\bibitem{goyal2018graph}
P.~Goyal and E.~Ferrara, ``Graph embedding techniques, applications, and
  performance: A survey,'' \emph{Knowledge-Based Systems}, vol. 151, pp.
  78--94, 2018.

\bibitem{pei2015nonnegative}
Y.~Pei, N.~Chakraborty, and K.~Sycara, ``Nonnegative matrix tri-factorization
  with graph regularization for community detection in social networks,'' in
  \emph{Twenty-Fourth International Joint Conference on Artificial
  Intelligence}, 2015.

\bibitem{wang2018billion}
J.~Wang, P.~Huang, H.~Zhao, Z.~Zhang, B.~Zhao, and D.~L. Lee, ``Billion-scale
  commodity embedding for e-commerce recommendation in alibaba,'' in
  \emph{Proceedings of the 24th ACM SIGKDD International Conference on
  Knowledge Discovery \& Data Mining}.\hskip 1em plus 0.5em minus 0.4em\relax
  ACM, 2018, pp. 839--848.

\bibitem{chen2018heterogeneous}
L.~Chen, Y.~Liu, Z.~Zheng, and P.~Yu, ``Heterogeneous neural attentive
  factorization machine for rating prediction,'' in \emph{CIKM}.\hskip 1em plus
  0.5em minus 0.4em\relax ACM, 2018, pp. 833--842.

\bibitem{xie2018factorization}
F.~Xie, L.~Chen, Y.~Ye, Z.~Zheng, and X.~Lin, ``Factorization machine based
  service recommendation on heterogeneous information networks,'' in
  \emph{ICWS}.\hskip 1em plus 0.5em minus 0.4em\relax IEEE, 2018, pp. 115--122.

\bibitem{zugner2018adversarial}
D.~Z{\"u}gner, A.~Akbarnejad, and S.~G{\"u}nnemann, ``Adversarial attacks on
  neural networks for graph data,'' in \emph{Proceedings of the 24th ACM SIGKDD
  International Conference on Knowledge Discovery \& Data Mining}.\hskip 1em
  plus 0.5em minus 0.4em\relax ACM, 2018, pp. 2847--2856.

\bibitem{zugner2019certifiable}
D.~Z{\"u}gner and S.~G{\"u}nnemann, ``Certifiable robustness and robust
  training for graph convolutional networks,'' in \emph{Proceedings of the 25th
  ACM SIGKDD International Conference on Knowledge Discovery \& Data
  Mining}.\hskip 1em plus 0.5em minus 0.4em\relax ACM, 2019, pp. 246--256.

\bibitem{wang2019adversarial}
S.~Wang, Z.~Chen, J.~Ni, X.~Yu, Z.~Li, H.~Chen, and P.~S. Yu, ``Adversarial
  defense framework for graph neural network,'' \emph{arXiv preprint
  arXiv:1905.03679}, 2019.

\bibitem{sun2018adversarial}
L.~Sun, J.~Wang, P.~S. Yu, and B.~Li, ``Adversarial attack and defense on graph
  data: A survey,'' \emph{arXiv preprint arXiv:1812.10528}, 2018.

\bibitem{ma2019attacking}
Y.~Ma, S.~Wang, L.~Wu, and J.~Tang, ``Attacking graph convolutional networks
  via rewiring,'' \emph{arXiv preprint arXiv:1906.03750}, 2019.

\bibitem{dai2018adversarial}
H.~Dai, H.~Li, T.~Tian, X.~Huang, L.~Wang, J.~Zhu, and L.~Song, ``Adversarial
  attack on graph structured data,'' \emph{arXiv preprint arXiv:1806.02371},
  2018.

\bibitem{chen2018link}
J.~Chen, Z.~Shi, Y.~Wu, X.~Xu, and H.~Zheng, ``Link prediction adversarial
  attack,'' \emph{arXiv preprint arXiv:1810.01110}, 2018.

\bibitem{chen2019can}
J.~Chen, Y.~Wu, X.~Lin, and Q.~Xuan, ``Can adversarial network attack be
  defended?'' \emph{arXiv preprint arXiv:1903.05994}, 2019.

\bibitem{chartrand1977introductory}
G.~Chartrand, \emph{Introductory graph theory}.\hskip 1em plus 0.5em minus
  0.4em\relax Courier Corporation, 1977.

\bibitem{ding2019deep}
K.~Ding, J.~Li, R.~Bhanushali, and H.~Liu, ``Deep anomaly detection on
  attributed networks,'' in \emph{Proceedings of the 2019 SIAM International
  Conference on Data Mining}.\hskip 1em plus 0.5em minus 0.4em\relax SIAM,
  2019, pp. 594--602.

\bibitem{liu2018heterogeneous}
Z.~Liu, C.~Chen, X.~Yang, J.~Zhou, X.~Li, and L.~Song, ``Heterogeneous graph
  neural networks for malicious account detection,'' in \emph{Proceedings of
  the 27th ACM International Conference on Information and Knowledge
  Management}.\hskip 1em plus 0.5em minus 0.4em\relax ACM, 2018, pp.
  2077--2085.

\bibitem{hussein2018meta}
R.~Hussein, D.~Yang, and P.~Cudr{\'e}-Mauroux, ``Are meta-paths necessary?:
  Revisiting heterogeneous graph embeddings,'' in \emph{Proceedings of the 27th
  ACM International Conference on Information and Knowledge Management}.\hskip
  1em plus 0.5em minus 0.4em\relax ACM, 2018, pp. 437--446.

\bibitem{manessi2020dynamic}
F.~Manessi, A.~Rozza, and M.~Manzo, ``Dynamic graph convolutional networks,''
  \emph{Pattern Recognition}, vol.~97, p. 107000, 2020.

\bibitem{kipf2017semi}
T.~N. Kipf and M.~Welling, ``Semi-supervised classification with graph
  convolutional networks,'' in \emph{International Conference on Learning
  Representations (ICLR)}, 2017.

\bibitem{sharma2018gradient}
Y.~Sharma, ``Gradient-based adversarial attacks to deep neural networks in
  limited access settings,'' Ph.D. dissertation, COOPER UNION, 2018.

\bibitem{chen2017practical}
Y.~Chen, Y.~Nadji, A.~Kountouras, F.~Monrose, R.~Perdisci, M.~Antonakakis, and
  N.~Vasiloglou, ``Practical attacks against graph-based clustering,'' in
  \emph{Proceedings of the 2017 ACM SIGSAC Conference on Computer and
  Communications Security}.\hskip 1em plus 0.5em minus 0.4em\relax ACM, 2017,
  pp. 1125--1142.

\bibitem{biggio2018wild}
B.~Biggio and F.~Roli, ``Wild patterns: Ten years after the rise of adversarial
  machine learning,'' \emph{Pattern Recognition}, vol.~84, pp. 317--331, 2018.

\bibitem{chen2019data}
L.~Chen, Y.~Xu, F.~Xie, M.~Huang, and Z.~Zheng, ``Data poisoning attacks on
  neighborhood-based recommender systems,'' 2019.

\bibitem{hou2019alphacyber}
S.~Hou, Y.~Fan, Y.~Zhang, Y.~Ye, J.~Lei, W.~Wan, J.~Wang, Q.~Xiong, and
  F.~Shao, ``$\alpha$cyber: Enhancing robustness of android malware detection
  system against adversarial attacks on heterogeneous graph based model,'' in
  \emph{Proceedings of the 28th ACM International Conference on Information and
  Knowledge Management}.\hskip 1em plus 0.5em minus 0.4em\relax ACM, 2019, pp.
  609--618.

\bibitem{DBLP:conf/ijcai/ChenL0GZ19}
L.~Chen, Y.~Liu, X.~He, L.~Gao, and Z.~Zheng, ``Matching user with item set:
  Collaborative bundle recommendation with deep attention network,'' in
  \emph{IJCAI}, 2019, pp. 2095--2101.

\bibitem{waniek2018attack}
M.~Waniek, K.~Zhou, Y.~Vorobeychik, E.~Moro, T.~P. Michalak, and T.~Rahwan,
  ``Attack tolerance of link prediction algorithms: How to hide your relations
  in a social network,'' \emph{arXiv preprint arXiv:1809.00152}, 2018.

\bibitem{bojchevski2019adversarial}
A.~Bojchevski and S.~G{\"u}nnemann, ``Adversarial attacks on node embeddings
  via graph poisoning,'' in \emph{International Conference on Machine
  Learning}, 2019, pp. 695--704.

\bibitem{xu2019topology}
K.~Xu, H.~Chen, S.~Liu, P.-Y. Chen, T.-W. Weng, M.~Hong, and X.~Lin, ``Topology
  attack and defense for graph neural networks: An optimization perspective,''
  \emph{arXiv preprint arXiv:1906.04214}, 2019.

\bibitem{zugner2019adversarial}
D.~Z{\"u}gner and S.~G{\"u}nnemann, ``Adversarial attacks on graph neural
  networks via meta learning,'' \emph{arXiv preprint arXiv:1902.08412}, 2019.

\bibitem{wang2018attack}
X.~Wang, J.~Eaton, C.-J. Hsieh, and F.~Wu, ``Attack graph convolutional
  networks by adding fake nodes,'' \emph{arXiv preprint arXiv:1810.10751},
  2018.

\bibitem{chen2018fast}
J.~Chen, Y.~Wu, X.~Xu, Y.~Chen, H.~Zheng, and Q.~Xuan, ``Fast gradient attack
  on network embedding,'' \emph{arXiv preprint arXiv:1809.02797}, 2018.

\bibitem{xuan2019unsupervised}
Q.~Xuan, J.~Zheng, L.~Chen, S.~Yu, J.~Chen, D.~Zhang, and Q.~Z. Member,
  ``Unsupervised euclidean distance attack on network embedding,'' \emph{arXiv
  preprint arXiv:1905.11015}, 2019.

\bibitem{bose2019generalizable}
A.~J. Bose, A.~Cianflone, and W.~Hamiltion, ``Generalizable adversarial attacks
  using generative models,'' \emph{arXiv preprint arXiv:1905.10864}, 2019.

\bibitem{wang2019attacking}
B.~Wang and N.~Z. Gong, ``Attacking graph-based classification via manipulating
  the graph structure,'' \emph{arXiv preprint arXiv:1903.00553}, 2019.

\bibitem{sun2018data}
M.~Sun, J.~Tang, H.~Li, B.~Li, C.~Xiao, Y.~Chen, and D.~Song, ``Data poisoning
  attack against unsupervised node embedding methods,'' \emph{arXiv preprint
  arXiv:1810.12881}, 2018.

\bibitem{zhou2019attacking}
K.~Zhou, T.~P. Michalak, M.~Waniek, T.~Rahwan, and Y.~Vorobeychik, ``Attacking
  similarity-based link prediction in social networks,'' in \emph{Proceedings
  of the 18th International Conference on Autonomous Agents and MultiAgent
  Systems}.\hskip 1em plus 0.5em minus 0.4em\relax International Foundation for
  Autonomous Agents and Multiagent Systems, 2018, pp. 305--313.

\bibitem{wu2019adversarial}
H.~Wu, C.~Wang, Y.~Tyshetskiy, A.~Docherty, K.~Lu, and L.~Zhu, ``Adversarial
  examples for graph data: deep insights into attack and defense,'' in
  \emph{Proceedings of the 28th International Joint Conference on Artificial
  Intelligence}.\hskip 1em plus 0.5em minus 0.4em\relax AAAI Press, 2019, pp.
  4816--4823.

\bibitem{perozzi2014deepwalk}
B.~Perozzi, R.~Al-Rfou, and S.~Skiena, ``Deepwalk: Online learning of social
  representations,'' in \emph{KDD}.\hskip 1em plus 0.5em minus 0.4em\relax ACM,
  2014, pp. 701--710.

\bibitem{christakopoulou2018adversarial}
K.~Christakopoulou and A.~Banerjee, ``Adversarial recommendation: Attack of the
  learned fake users,'' \emph{arXiv preprint arXiv:1809.08336}, 2018.

\bibitem{chen2019time}
J.~Chen, J.~Zhang, Z.~Chen, M.~Du, F.~Li, and Q.~Xuan, ``Time-aware gradient
  attack on dynamic network link prediction,'' \emph{arXiv preprint
  arXiv:1911.10561}, 2019.

\bibitem{chen2019ga}
J.~Chen, L.~Chen, Y.~Chen, M.~Zhao, S.~Yu, Q.~Xuan, and X.~Yang, ``Ga-based
  q-attack on community detection,'' \emph{IEEE Transactions on Computational
  Social Systems}, vol.~6, no.~3, pp. 491--503, 2019.

\bibitem{chang2019restricted}
H.~Chang, Y.~Rong, T.~Xu, W.~Huang, H.~Zhang, P.~Cui, W.~Zhu, and J.~Huang, ``A
  restricted black-box adversarial framework towards attacking graph embedding
  models,'' 2019.

\bibitem{chen2019multiscale}
J.~Chen, Y.~Chen, L.~Chen, M.~Zhao, and Q.~Xuan, ``Multiscale evolutionary
  perturbation attack on community detection,'' \emph{arXiv preprint
  arXiv:1910.09741}, 2019.

\bibitem{zhang2019data}
H.~Zhang, T.~Zheng, J.~Gao, C.~Miao, L.~Su, Y.~Li, and K.~Ren, ``Data poisoning
  attack against knowledge graph embedding,'' in \emph{Proceedings of the 28th
  International Joint Conference on Artificial Intelligence}.\hskip 1em plus
  0.5em minus 0.4em\relax AAAI Press, 2019, pp. 4853--4859.

\bibitem{whitley1994genetic}
D.~Whitley, ``A genetic algorithm tutorial,'' \emph{Statistics and computing},
  vol.~4, no.~2, pp. 65--85, 1994.

\bibitem{sutton2018reinforcement}
R.~S. Sutton and A.~G. Barto, \emph{Reinforcement learning: An
  introduction}.\hskip 1em plus 0.5em minus 0.4em\relax MIT press, 2018.

\bibitem{goodfellow2014generative}
I.~J. Goodfellow, J.~Pouget-Abadie, M.~Mirza, B.~Xu, D.~Warde-Farley, S.~Ozair,
  A.~Courville, and Y.~Bengio, ``Generative adversarial networks,'' 2014.

\bibitem{zhu2016gemini}
X.~Zhu, W.~Chen, W.~Zheng, and X.~Ma, ``Gemini: A computation-centric
  distributed graph processing system,'' in \emph{12th $\{$USENIX$\}$ Symposium
  on Operating Systems Design and Implementation ($\{$OSDI$\}$ 16)}, 2016, pp.
  301--316.

\bibitem{yang2019knightking}
K.~Yang, M.~Zhang, K.~Chen, X.~Ma, Y.~Bai, and Y.~Jiang, ``Knightking: a fast
  distributed graph random walk engine,'' in \emph{Proceedings of the 27th ACM
  Symposium on Operating Systems Principles}, 2019, pp. 524--537.

\bibitem{zhu2019robust}
D.~Zhu, Z.~Zhang, P.~Cui, and W.~Zhu, ``Robust graph convolutional networks
  against adversarial attacks,'' 2019.

\bibitem{tang2019robust}
X.~Tang, Y.~Li, Y.~Sun, H.~Yao, P.~Mitra, and S.~Wang, ``Robust graph neural
  network against poisoning attacks via transfer learning,'' \emph{arXiv
  preprint arXiv:1908.07558}, 2019.

\bibitem{ioannidis2019edge}
V.~N. Ioannidis and G.~B. Giannakis, ``Edge dithering for robust adaptive graph
  convolutional networks,'' \emph{arXiv preprint arXiv:1910.09590}, 2019.

\bibitem{sun2019virtual}
K.~Sun, Z.~Lin, H.~Guo, and Z.~Zhu, ``Virtual adversarial training on graph
  convolutional networks in node classification,'' in \emph{Chinese Conference
  on Pattern Recognition and Computer Vision (PRCV)}.\hskip 1em plus 0.5em
  minus 0.4em\relax Springer, 2019, pp. 431--443.

\bibitem{feng2019graph}
F.~Feng, X.~He, J.~Tang, and T.-S. Chua, ``Graph adversarial training:
  Dynamically regularizing based on graph structure,'' \emph{arXiv preprint
  arXiv:1902.08226}, 2019.

\bibitem{wang2019graphdefense}
X.~Wang, X.~Liu, and C.-J. Hsieh, ``Graphdefense: Towards robust graph
  convolutional networks,'' \emph{arXiv preprint arXiv:1911.04429}, 2019.

\bibitem{deng2019batch}
Z.~Deng, Y.~Dong, and J.~Zhu, ``Batch virtual adversarial training for graph
  convolutional networks,'' \emph{arXiv preprint arXiv:1902.09192}, 2019.

\bibitem{he2018adversarial}
X.~He, Z.~He, X.~Du, and T.-S. Chua, ``Adversarial personalized ranking for
  recommendation,'' in \emph{The 41st International ACM SIGIR Conference on
  Research \& Development in Information Retrieval}.\hskip 1em plus 0.5em minus
  0.4em\relax ACM, 2018, pp. 355--364.

\bibitem{bojchevski2019certifiable}
A.~Bojchevski and S.~G{\"u}nnemann, ``Certifiable robustness to graph
  perturbations,'' in \emph{Advances in Neural Information Processing Systems},
  2019, pp. 8317--8328.

\bibitem{jin2019power}
M.~Jin, H.~Chang, W.~Zhu, and S.~Sojoudi, ``Power up! robust graph
  convolutional network against evasion attacks based on graph powering,''
  \emph{arXiv preprint arXiv:1905.10029}, 2019.

\bibitem{pezeshkpour2019investigating}
P.~Pezeshkpour, C.~Irvine, Y.~Tian, and S.~Singh, ``Investigating robustness
  and interpretability of link prediction via adversarial modifications,'' in
  \emph{Proceedings of NAACL-HLT}, 2019, pp. 3336--3347.

\bibitem{xu2018characterizing}
X.~Xu, Y.~Yu, B.~Li, L.~Song, C.~Liu, and C.~Gunter, ``Characterizing malicious
  edges targeting on graph neural networks,'' 2018.

\bibitem{zhang2019comparing}
Y.~Zhang, S.~Khan, and M.~Coates, ``Comparing and detecting adversarial attacks
  for graph deep learning,'' in \emph{Proc. Representation Learning on Graphs
  and Manifolds Workshop, Int. Conf. Learning Representations, New Orleans, LA,
  USA}, 2019.

\bibitem{sun2011pathsim}
Y.~Sun, J.~Han, X.~Yan, P.~S. Yu, and T.~Wu, ``Pathsim: Meta path-based top-k
  similarity search in heterogeneous information networks,'' \emph{Proceedings
  of the VLDB Endowment}, vol.~4, no.~11, pp. 992--1003, 2011.

\bibitem{ioannidis2019graphsac}
V.~N. Ioannidis, D.~Berberidis, and G.~B. Giannakis, ``Graphsac: Detecting
  anomalies in large-scale graphs,'' \emph{arXiv preprint arXiv:1910.09589},
  2019.

\bibitem{miller2019improving}
B.~A. Miller, M.~{\c{C}}amurcu, A.~J. Gomez, K.~Chan, and T.~Eliassi-Rad,
  ``Improving robustness to attacks against vertex classification,'' 2019.

\bibitem{zhou2019adversarial}
K.~Zhou, T.~P. Michalak, and Y.~Vorobeychik, ``Adversarial robustness of
  similarity-based link prediction,'' \emph{arXiv preprint arXiv:1909.01432},
  2019.

\bibitem{goodfellow2014explaining}
I.~J. Goodfellow, J.~Shlens, and C.~Szegedy, ``Explaining and harnessing
  adversarial examples,'' \emph{arXiv preprint arXiv:1412.6572}, 2014.

\bibitem{rendle2009bpr}
S.~Rendle, C.~Freudenthaler, Z.~Gantner, and L.~Schmidt-Thieme, ``Bpr: Bayesian
  personalized ranking from implicit feedback,'' in \emph{Proceedings of the
  twenty-fifth conference on uncertainty in artificial intelligence}.\hskip 1em
  plus 0.5em minus 0.4em\relax AUAI Press, 2009, pp. 452--461.

\bibitem{vaswani2017attention}
A.~Vaswani, N.~Shazeer, N.~Parmar, J.~Uszkoreit, L.~Jones, A.~N. Gomez,
  {\L}.~Kaiser, and I.~Polosukhin, ``Attention is all you need,'' in
  \emph{Advances in neural information processing systems}, 2017, pp.
  5998--6008.

\bibitem{kullback1951information}
S.~Kullback and R.~A. Leibler, ``On information and sufficiency,'' \emph{The
  annals of mathematical statistics}, vol.~22, no.~1, pp. 79--86, 1951.

\bibitem{von2010market}
H.~Von~Stackelberg, \emph{Market structure and equilibrium}.\hskip 1em plus
  0.5em minus 0.4em\relax Springer Science \& Business Media, 2010.

\bibitem{powers2011evaluation}
D.~M. Powers, ``Evaluation: from precision, recall and f-measure to roc,
  informedness, markedness and correlation,'' 2011.

\bibitem{boyd2013area}
K.~Boyd, K.~H. Eng, and C.~D. Page, ``Area under the precision-recall curve:
  point estimates and confidence intervals,'' in \emph{Joint European
  conference on machine learning and knowledge discovery in databases}.\hskip
  1em plus 0.5em minus 0.4em\relax Springer, 2013, pp. 451--466.

\bibitem{newman2003mixing}
M.~E. Newman, ``Mixing patterns in networks,'' \emph{Physical Review E},
  vol.~67, no.~2, p. 026126, 2003.

\bibitem{newman2004finding}
M.~E. Newman and M.~Girvan, ``Finding and evaluating community structure in
  networks,'' \emph{Physical review E}, vol.~69, no.~2, p. 026113, 2004.

\bibitem{danon2005comparing}
L.~Danon, A.~Diaz-Guilera, J.~Duch, and A.~Arenas, ``Comparing community
  structure identification,'' \emph{Journal of Statistical Mechanics: Theory
  and Experiment}, vol. 2005, no.~09, p. P09008, 2005.

\bibitem{santos2009use}
J.~M. Santos and M.~Embrechts, ``On the use of the adjusted rand index as a
  metric for evaluating supervised classification,'' in \emph{International
  conference on artificial neural networks}.\hskip 1em plus 0.5em minus
  0.4em\relax Springer, 2009, pp. 175--184.

\bibitem{lu2011link}
L.~L{\"u} and T.~Zhou, ``Link prediction in complex networks: A survey,''
  \emph{Physica A: statistical mechanics and its applications}, vol. 390,
  no.~6, pp. 1150--1170, 2011.

\bibitem{mccallum2000automating}
A.~K. McCallum, K.~Nigam, J.~Rennie, and K.~Seymore, ``Automating the
  construction of internet portals with machine learning,'' \emph{Information
  Retrieval}, vol.~3, no.~2, pp. 127--163, 2000.

\bibitem{sen2008collective}
P.~Sen, G.~Namata, M.~Bilgic, L.~Getoor, B.~Galligher, and T.~Eliassi-Rad,
  ``Collective classification in network data,'' \emph{AI magazine}, vol.~29,
  no.~3, pp. 93--93, 2008.

\bibitem{madry2017towards}
A.~Madry, A.~Makelov, L.~Schmidt, D.~Tsipras, and A.~Vladu, ``Towards deep
  learning models resistant to adversarial attacks,'' \emph{arXiv preprint
  arXiv:1706.06083}, 2017.

\end{thebibliography}
\end{document}